\documentclass{article}

\usepackage{times}

\usepackage{graphicx} 
\usepackage{subfigure}

\usepackage{natbib}

\usepackage{appendix}

\usepackage[left=1.25in,top=1.25in,right=1.25in,bottom=1.25in,head=1.25in]{geometry}
\usepackage[english]{babel}
\usepackage{amsfonts,amsmath,amssymb,amsthm}
\usepackage{dsfont,cancel}

\usepackage{algorithm}
\usepackage{algorithmic}

\usepackage{hyperref}
\numberwithin{equation}{section}
\numberwithin{figure}{section}

\newtheorem{theorem}{Theorem}
\newtheorem{lemma}{Lemma}
\newtheorem{corollary}{Corollary}

\theoremstyle{remark}

\theoremstyle{definition}

\newcommand{\argmin}{\mathop{\mathrm{argmin}}}

\def\E{\mathbb{E}}
\def\P{\mathbb{P}}

\def\half{\frac{1}{2}}

\def\diag{\mathrm{diag}}

\def\hbeta{\hat{\beta}}

\def\halpha{\hat{\alpha}}
\def\hf{\hat{f}}

\def\R{\mathbb{R}}

\def\cG{\mathcal{G}}
\def\cH{\mathcal{H}}

\def\cS{\mathcal{S}}

\def\TV{\mathrm{TV}}
\def\KS{\mathrm{KS}}

\begin{document}

\title{The Falling Factorial Basis and Its Statistical Applications}

%

\author{
Yu-Xiang Wang$^{*}$ \\
{\tt \normalsize yuxiangw@cs.cmu.edu} \\
\and
Alex Smola$^{*}$ \\
{\tt \normalsize alex@smola.org} \\
\and
Ryan J. Tibshirani$^{\dag,*}$\\
{\tt \normalsize ryantibs@stat.cmu.edu}\\
\and
  \begin{tabular}{c}
    Machine Learning Department$^{*}$, Department of Statisics$^{\dag}$, \\
    Carnegie Mellon University, Pittsburgh, PA 15213
  \end{tabular}
}

\date{}

\maketitle
\begin{abstract}
We study a novel spline-like basis, which we name the
``falling factorial basis'', bearing many similarities to the
classic truncated power basis.  The advantage of the falling factorial
basis is that it enables rapid, linear-time computations in basis
matrix multiplication and basis matrix inversion.  The falling
factorial functions are not actually splines, but are close enough
to splines that they provably retain some of the favorable properties
of the latter functions.  We examine their application in two
problems: trend filtering over arbitrary input points, and a
higher-order variant of the two-sample Kolmogorov-Smirnov test.
\end{abstract}

\section{Introduction}
\label{sec:intro}

Splines are an old concept, and they play important roles in various
subfields of mathematics and statistics; see e.g.,
\citet{deboorsplines}, \citet{wahbasplines} for two classic
references.
In words, a spline of order $k$ is a piecewise polynomial
of degree $k$ that is continuous and has continuous derivatives of
orders $1,2,\ldots k-1$ at its knot points.  In this paper, we
look at a new twist on an old problem: we examine a novel set of
spline-like basis functions with sound computational and statistical
properties.  This basis, which we call the
{\it falling factorial basis},
is particularly attractive when assessing higher
order of smoothness via the total variation operator, due to the
capability for sparse decompositions.
A summary of our main findings is as follows.
\begin{itemize}
\item The falling factorial basis and its inverse both admit a
  linear-time transformation, i.e., much faster decompositions than
  the spline basis, and even faster than, e.g., the fast Fourier
  transform.
\item For all practical purposes, the falling factorial basis shares
  the statistical properties of the spline basis. We derive a
  sharp characterization of the discrepancy between the two bases in
  terms of the polynomial degree and the distance between sampling
  points.

\item We simplify and extend known convergence results on trend
  filtering, a nonparametric regression technique that implicitly
  employs the falling factorial basis.

\item We also extend the Kolmogorov-Smirnov two-sample test to
  account for higher order differences, and utilize the falling
  factorial basis for rapid computations.  We provide no theory
  but demonstrate excellent empirical results, improving on, e.g., the
  maximum mean discrepancy \citep{gretton2012kernel} and
  Anderson-Darling \citep{anderson1954test} tests.
\end{itemize}
In short, the falling factorial function class offers an exciting
prospect for univariate function regularization.

Now let us review some basics. Recall that
the set of $k$th order splines with knots over a fixed set
of $n$ points forms an $(n+k+1)$-dimensional subspace of functions.
Here and throughout, we assume that we are given
ordered input points $x_1 < x_2 < \ldots < x_n$ and a
polynomial order $k \geq 0$, and we
define a set of knots $T=\{t_1,\ldots t_{n-k-1}\}$ by excluding some
of the input points at the left and right boundaries, in particular,
\begin{equation}
\label{eq:t}
T = \begin{cases}
\{x_{k/2+2},\ldots x_{n-k/2}\} & \text{if $k$ is even}, \\
\{x_{(k+1)/2+1},\ldots x_{n-(k+1)/2}\} & \text{if $k$ is odd}.
\end{cases}
\end{equation}
The set of $k$th order splines with knots in $T$ hence forms an
$n$-dimensional subspace of functions.  The canonical parametrization
for this subspace is given by the truncated power basis, $g_1,\ldots
g_n$, defined as
\begin{equation}
\label{eq:gbasis}
\begin{gathered}
g_1(x)=1, \; g_2(x)=x, \;\ldots\;
g_{k+1}(x) = x^k, \\
g_{k+1+j}(x) = (x-t_j)^k \cdot
1\{x \geq t_j\}, \;\;
\hfill j=1,\ldots n-k-1.
\end{gathered}
\end{equation}
These functions can also be used to define the truncated power
basis matrix, $G \in \R^{n\times n}$, by
\begin{equation}
\label{eq:g}
G_{ij} = g_j(x_i), \;\;\; i,j=1,\ldots n,
\end{equation}
i.e., the columns of $G$ give the evaluations of the basis functions
$g_1,\ldots g_n$ over the inputs $x_1,\ldots x_n$.
As $g_1,\ldots g_n$ are linearly independent functions, $G$ has
linearly independent columns, and hence $G$ is invertible.

As noted, our focus is a related but different set of
basis functions, named the falling factorial basis functions. We
define these functions, for a given order $k\geq 0$, as
\begin{equation}
\label{eq:hbasis}
\begin{gathered}
h_j(x) = \prod_{\ell=1}^{j-1} (x-x_\ell), \;\;\;
j=1,\ldots k+1,\\
h_{k+1+j}(x) = \prod_{\ell=1}^k (x-x_{j+\ell}) \cdot 1\{x \geq
x_{j+k}\}, \;\;
\hfill  j=1,\ldots n-k-1.
\end{gathered}
\end{equation}
(Our convention is to take the empty product to be 1, so that
$h_1(x)=1$.)  The falling factorial basis functions are piecewise
polynomial, and have an analogous form to the truncated power basis
functions in \eqref{eq:gbasis}.  Loosely speaking, they are given by
replacing an $r$th order power function in the truncated power basis
with an appropriate $r$-term product, e.g., replacing $x^2$ with
$(x-x_2)(x-x_1)$, and $(x-t_j)^k$ with
$(x-x_{j+k})(x-x_{j+k-1})\cdot \ldots (x-x_{j+1})$.
Similar to the above, we can define the falling factorial basis matrix,
$H \in \R^{n\times n}$, by
\begin{equation}
\label{eq:h}
H_{ij} = h_j(x_i), \;\;\; i,j=1,\ldots n,
\end{equation}
and the linear independence of $h_1,\ldots h_n$ implies that $H$ too
is invertible.

Note that the first $k+1$ functions of either basis, the truncated
power or falling factorial basis, span the same
space (the space of $k$th order polynomials).  But this is not
true of the last $n-k-1$ functions.
Direct calculation shows that, while continuous, the function
$h_{j+k+1}$ has discontinuous derivatives of all orders $1,\ldots k$
at the point $x_{j+k}$, for $j=1,\ldots n-k-1$.
This means that the falling factorial functions $h_{k+2},\ldots h_n$
are not actually $k$th order splines, but are instead continuous
$k$th order piecewise polynomials that are ``close to'' splines.
Why would we ever use such a seemingly strange basis as that
defined in \eqref{eq:hbasis}? To repeat what was summarized above,
the falling factorial functions allow for linear-time (and
closed-form) computations with the basis matrix $H$ and its
inverse. Meanwhile, the falling factorial functions are close enough
to the truncated power functions that using them in several
spline-based problems (i.e., using $H$ in place of $G$) can be
statistically legitimized.
We make this statement precise in the sections that follow.

As we see it, there is really nothing about their form in
\eqref{eq:hbasis} that suggests a particularly special computational
structure of the falling factorial basis functions.  Our interest in
these functions arose from a study of trend filtering, a
nonparametric regression estimator, where the inverse of $H$ plays a
natural role.  The inverse of $H$ 
is a kind of discrete derivative
operator of order $k+1$, properly adjusted for the spacings between
the input points $x_1,\ldots x_n$.  It is really the special, banded
structure of this derivative operator that underlies the
computational efficiency surrounding the falling factorial basis; all
of the computational routines proposed in this paper leverage this
structure.

Here is an outline for rest of this article.
In Section \ref{sec:props}, we describe a number of basic properties
of the falling factorial basis functions, culminating in fast
linear-time algorithms for multiplication $H$ and $H^{-1}$,
and tight error bounds between $H$ and the truncated power basis
matrix $G$. Section \ref{sec:bsplines} discusses B-splines, which
provide another highly
efficient basis for spline manipulations; we explain why the
falling factorial basis offers a preferred parametrization in some
specific statistical applications, e.g., the ones we present in
Sections \ref{sec:tf} and \ref{sec:ks}.  Section \ref{sec:tf}
covers trend filtering, and extends a known convergence result
for trend filtering over evenly spaced input
points \citep{trendfilter} to the case of arbitrary input
points. The conclusion is that trend filtering estimates converge at
the minimax rate (over a large class of true functions) assuming only
mild conditions on the inputs.  In Section \ref{sec:ks}, we consider a
higher order extension of the classic two-sample Kolmogorov-Smirnov
test.  We find this test to have better power in detecting higher
order (tail) differences between distributions when compared to the
usual Kolmogorov-Smirnov test; furthermore, by employing the falling
factorial functions, it can computed in linear time.  In Section
\ref{sec:discuss}, we end with some discussion.

\section{Basic properties}
\label{sec:props}

Consider the falling factorial basis matrix $H \in \R^{n\times n}$,
as defined in \eqref{eq:h}, over input points $x_1 < \ldots < x_n$.
The following subsections describe a recursive decomposition for $H$
and its inverse, which lead to fast computational methods for
multiplication by $H$ and $H^{-1}$ (as well as $H^T$ and
$(H^T)^{-1}$). The last subsection bounds
the maximum absolute difference bewteen the elements of $H$
and $G$, the truncated power basis matrix (also defined over
$x_1, \ldots x_n$).
Lemmas \ref{lem:recursive},
\ref{lem:hinv}, \ref{lem:hgdiff} below were derived in
\citet{trendfilter} for the special case of evenly spaced inputs,
$x_i=i/n$ for $i=1,\ldots n$.  We reiterate that here we
consider generic input points $x_1,\ldots x_n$.
In the interest of space, we defer all proofs to the appendix.

\subsection{Recursive decomposition}
\label{sec:recursive}

Our first result shows that $H$ decomposes into a product of simpler
matrices. It helpful to define, for $k \geq 1$,
\begin{equation*}
\Delta^{(k)}=\diag
\big(x_{k+1}-x_1, \, x_{k+2}-x_2, \, \ldots \,
x_n-x_{n-k}\big),
\end{equation*}
the $(n-k)\times (n-k)$ diagonal matrix whose
diagonal elements contain the $k$-hop gaps between input points.
\begin{lemma}
\label{lem:recursive}
Let $I_m$ denote the $m\times m$ identity matrix, and $L_m$
the $m\times m$ lower triangular matrix of 1s.
If we write $H^{(k)}$ for the falling factorial basis matrix of
order $k$, then in this notation, we have $H^{(0)}=L_n$, and for
$k \geq 1$,
\begin{equation}
\label{eq:hk}
H^{(k)} = H^{(k-1)} \cdot
\left[\begin{array}{cc}
I_k & 0 \\ 0 & \Delta^{(k)}L_{n-k}
\end{array}\right].
\end{equation}

\end{lemma}

Lemma \ref{lem:recursive} is really a key workhorse behind many
properties of the falling factorial basis functions.
E.g., it acts as a building block for
results to come: immediately, the representation \eqref{eq:hk}
suggests both an analogous inverse representation for $H^{(k)}$, and a
computational strategy for matrix multiplication by $H^{(k)}$.  These
are discussed in the next two subsections.
We remark that the result in the lemma
may seem surprising, as there is not an apparent
connection between the falling factorial functions in
\eqref{eq:hbasis} and the recursion in \eqref{eq:hk}, which is based
on taking cumulative sums at varying offsets
(the rightmost matrix in
\eqref{eq:hk}).  We were led to this result by studying the evenly
spaced case; its proof for the present case
is considerably longer and more technical, but the statement of the
lemma is still quite simple.

\subsection{The inverse basis}

The result in Lemma \ref{lem:recursive} clearly also implies a result on
the inverse operators, namely, that $(H^{(0)})^{-1} = L_n^{-1}$, and
\begin{equation}
\label{eq:hkinv}
(H^{(k)})^{-1} =
\left[\begin{array}{cc}
I_k & 0 \\ 0 & L_{n-k}^{-1}(\Delta^{(k)})^{-1}
\end{array}\right]
\cdot
(H^{(k-1)})^{-1}
\end{equation}
for all $k \geq 1$. We note that
\begin{equation}
\label{eq:linv}
L_m^{-1} = \left[\begin{array}{cccc}
1 & 0 & \ldots & 0 \\
1 & 1 & \ldots & 0 \\
\vdots & & &  \\
1 & 1 & \ldots & 1
\end{array}\right]^{-1} =
\left[\begin{array}{c} e_1^T \\ D^{(1)}
\end{array}\right],
\end{equation}
with $e_1 = (1,0,\ldots 0) \in \R^m$ being the first standard basis
vector, and $D^{(1)} \in \R^{(m-1)\times m}$ the first discrete
difference operator
\begin{equation}
\label{eq:d1}
D^{(1)} =
\left[\begin{array}{rrrrrr}
-1 & 1 & 0 & \ldots & 0 & 0 \\
0 & -1 & 1 & \ldots & 0 & 0 \\
\vdots & & & & & \\
0 & 0 & 0 & \ldots & -1 & 1
\end{array}\right],
\end{equation}
With this in mind, the recursion in \eqref{eq:hkinv} now looks like
the construction of the higher order discrete difference operators,
over the input $x_1,\ldots x_n$.  To define these operators, we
start with the first order discrete difference operator
$D^{(1)} \in \R^{(n-1)\times n}$
as in \eqref{eq:d1}, and define the higher order difference
discrete operators according to
\begin{equation}
\label{eq:dk}
D^{(k+1)} = D^{(1)} \cdot
k\cdot (\Delta^{(k)})^{-1} \cdot
D^{(k)},
\end{equation}
for $k \geq 1$.  As $D^{(k+1)} \in \R^{(n-k-1)\times n}$,
leading matrix $D^{(1)}$ above denotes the $(n-k-1) \times
(n-k)$ version of the first order difference operator in \eqref{eq:d1}.

To gather intuition, we can think of $D^{(k)}$ as a type of
discrete $k$th order derivative operator across the
underlying points $x_1,\ldots x_n$; i.e., given an arbitrary
sequence $u=(u_1,\ldots u_n)\in\R^n$ over the positions
$x_1,\ldots x_n$, respectively, we can think of $(D^{(k)} u)_i$ as the
discrete $k$th derivative of the sequence $u$ evaluated at the point
$x_i$. It is not difficult to see, from its definition, that $D^{(k)}$ is a
banded matrix with bandwidth $k+1$.  The middle (diagonal) term in
\eqref{eq:dk} accounts for the fact that the underlying positions
$x_1,\ldots x_n$ are not necessarily evenly spaced.  When the input
points are evenly spaced, this term contributes only a constant
factor, and the difference operators $D^{(k)}$, $k=1,2,3,\ldots$ take
a very simple form,  where each row is a shifted version of the
previous, and the nonzero elements are given by the $k$th order
binomial coefficients (with alternating signs); see
\citet{trendfilter}.

By staring at \eqref{eq:hkinv} and \eqref{eq:dk}, one can see that the
falling factorial basis matrices and discrete difference operators are
essentially inverses of each other.  The story is only slightly more
complicated because the difference matrices are not square.

\begin{lemma}
\label{lem:hinv}
If $H^{(k)}$ is the $k$th order falling factorial basis matrix
defined over the inputs
$x_1, \ldots x_n$, and $D^{(k+1)}$ is the $(k+1)$st order discrete
difference operator defined over the same inputs
$x_1\ldots x_n$, then
\begin{equation}\label{eq:invH}
(H^{(k)})^{-1} = \left[\begin{array}{c} C \\
\frac{1}{k!} \cdot D^{(k+1)}
\end{array}\right],
\end{equation}
for an explicit matrix $C \in \R^{(k+1)\times n}$.  If we let $A_i$
denote the $i$th row of a matrix $A$, then $C$ has first
row $C_1 = e_1^T$, and subsequent rows
\begin{equation*}
C_{i+1} =
\left[ \frac{1}{(i-1)!} \cdot (\Delta^{(i)})^{-1} \cdot D^{(i)}\right]_1,
\;\;\; i=1,\ldots k.
\end{equation*}
\end{lemma}

Lemma \ref{lem:hinv} shows that the last $n-k-1$ rows
of $(H^{(k)})^{-1}$ are given exactly by $D^{(k+1)}/k!$.  This
serves as the crucial link between the falling factorial basis
functions and trend filtering, discussed in Section \ref{sec:tf}.  The
route to proving this result revealed the recursive expressions
\eqref{eq:hk} and \eqref{eq:hkinv}, and in fact these are of great
computational interest in their own right, as we discuss next.

\subsection{Fast matrix multiplication}
\label{sec:fastcomp}

The recursions in \eqref{eq:hk} and \eqref{eq:hkinv} allow us to apply
$H^{(k)}$ and $(H^{(k)})^{-1}$ with specialized linear-time algorithms.
Further, these algorithms are completely in-place: we do not
need to form the matrices $H^{(k)}$ or $(H^{(k)})^{-1}$, and the
algorithms operate entirely by manipulating the input vector (the
vector to be multiplied).

\begin{lemma}
\label{lem:fastcomp}
For the $k$th order falling factorial basis matrix
$H^{(k)} \in \R^{n\times n}$, over arbitrary sorted inputs $x_1,\ldots x_n$,
multiplication by $H^{(k)}$ and $(H^{(k)})^{-1}$ can each be
computed in $O(nk)$ in-place operations with zero memory requirements
(aside from storing the input points and the vector to be multiplied),
i.e., we do not need to form $H^{(k)}$ or $(H^{(k)})^{-1}$. Algorithms
\ref{alg:hk} and \ref{alg:hkinv} give the details.
The same is true for matrix multiplication by $(H^{(k)})^T$ and
$[(H^{(k)})^T]^{-1}$; Algorithms \ref{alg:hkt} and \ref{alg:hktinv},
found in the appendix, give the details.
\end{lemma}

\begin{algorithm}[tb]
\caption{Multiplication by $H^{(k)}$}
\label{alg:hk}
\begin{algorithmic}
\STATE {\bfseries Input:} Vector to be multiplied $y \in \R^n$,
order $k \geq 0$, sorted inputs vector $x\in \R^n$.
\STATE {\bfseries Output:} $y$ is overwritten by $H^{(k)}y$.
\FOR{$i=k$ to $0$}
\STATE $y_{(i+1):n}=\mathrm{cumsum}(y_{(i+1):n})$, \\
where $y_{a:b}$ denotes the subvector $(y_a,y_{a+1},...,y_b)$ and
$\mathrm{cumsum}$ is the cumulative sum operator.
\IF{$i\neq 0$}
\STATE $y_{(i+1):n}=(x_{(i+1):n}-x_{1:(n-i)})
\, .\hspace{-2pt} * \,
 y_{(i+1):n}$, \\
where $.*$ denotes entrywise multiplication.
\ENDIF
\ENDFOR
\STATE Return $y$.
\end{algorithmic}
\end{algorithm}

\begin{algorithm}[tb]
\caption{Multiplication by $(H^{(k)})^{-1}$}
\label{alg:hkinv}
\begin{algorithmic}
\STATE {\bfseries Input:} Vector to be multiplied $y \in \R^n$,
order $k \geq 0$, sorted inputs vector $x\in \R^n$.
\STATE {\bfseries Output:} $y$ is overwritten by $(H^{(k)})^{-1} y$.
\FOR{$i=0$ to $k$}
\IF{$i\neq 0$}
\STATE $y_{(i+1):n}=y_{i+1:n} \,./ \,(x_{(i+1):n}-x_{1:(n-i]})$, \\
where $./$ is entrywise division.
\ENDIF
\STATE $y_{(i+2):n} = \mathrm{diff}(y_{(i+1):n})$, \\
where $\mathrm{diff}$ is the pairwise difference operator.
\ENDFOR
\STATE Return $y$.
\end{algorithmic}
\end{algorithm}

Note that the lemma assumes presorted inputs $x_1,\ldots x_n$
(sorting requires an extra $O(n\log{n})$
operations). The routines for multiplication by $H^{(k)}$ and
$(H^{(k)})^{-1}$, in Algorithms \ref{alg:hk} and \ref{alg:hkinv},
are really just given by inverting each term one at a time in the
product representations \eqref{eq:hk} and \eqref{eq:hkinv}.
They are composed of elementary in-place
operations, like cumulative sums and pairwise differences.
This brings to mind a comparison to wavelets, as both the wavelet and
inverse wavelets operators can be viewed as highly specialized
linear-time matrix multplications.

Borrowing from the wavelet
perspective, given a sampled signal $y_i = f(x_i)$, $i=1,\ldots n$,
the action $(H^{(k)})^{-1}y$ can be thought of as the forward
transform under the piecewise polynomial falling factorial basis, and
$H^{(k)} y$ as the backward or inverse transform under this basis.
It might be interesting to consider the applicability of such
transforms to signal processing tasks, but this is beyond the scope
of the current paper, and we leave it to potential future work.

\begin{figure}[tb]
  \centering
\subfigure[\scriptsize{Falling factorial vs.\ Fourier, wavelet,
  and B-spline transforms (linear scale)}]{
  \includegraphics[width=0.4725\linewidth]{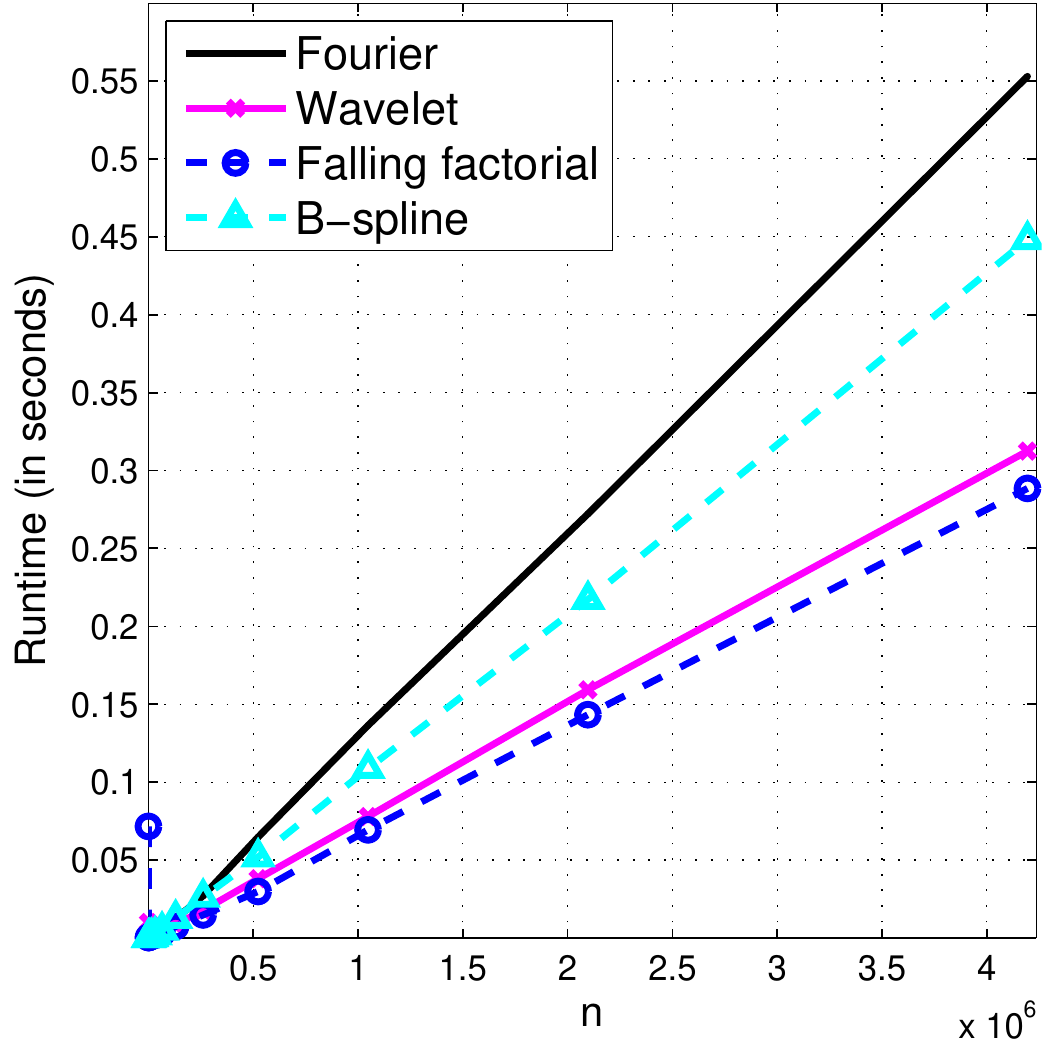}
  \label{fig:runtimes1}
}
\subfigure[\scriptsize{Falling factorial (H) vs.\ truncated power (G)
  transforms (log-log scale)}]{
  \includegraphics[width=0.4725\linewidth]{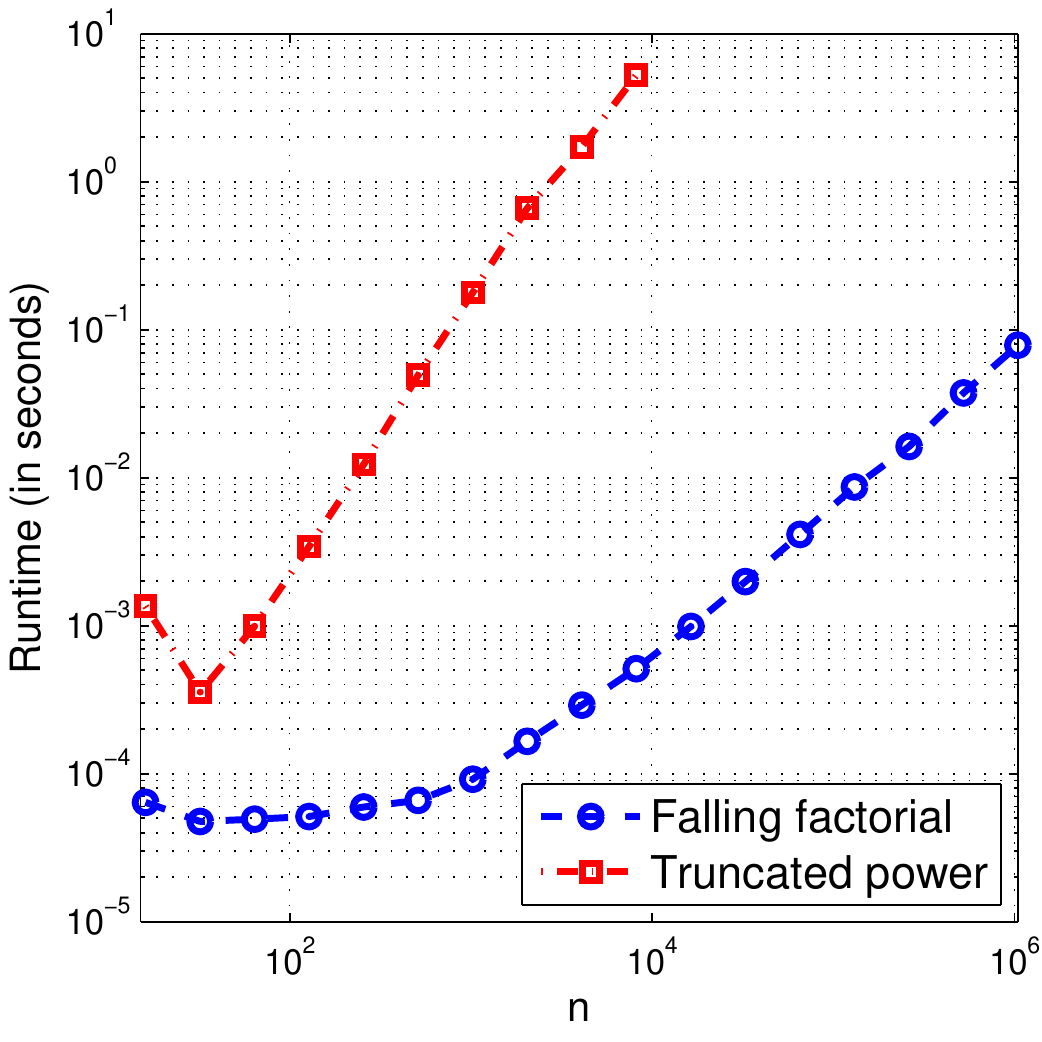}
  \label{fig:runtimes2}
}
\caption{Comparison of runtimes for different transforms. The
  experiments were performed on a laptop computer.}
\label{fig:runtime}
\end{figure}

We do however include a computational comparison between
the forward and backward falling factorial transforms, in Algorithms
\ref{alg:hkinv} and \ref{alg:hk}, and the well-studied Fourier and
wavelet transforms.  Figure \ref{fig:runtimes1} shows
the runtimes of one complete cycle of falling factorial
transforms (i.e., one forward and one backward transform),
with $k=3$, versus one cycle of fast Fourier transforms and one cycle
of wavelet transforms (using symmlets).
The comparison was run in Matlab, and we used Matlab's ``fft'' and
``ifft'' functions for the fast Fourier transforms, and the Stanford
WaveLab's ``FWT\_PO'' and ``IWT\_PO'' functions (with symmlet filters)
for the wavelet transforms \citep{buckheit1995wavelab}.
These functions all call on C implementations that have been ported
to Matlab using MEX-functions, and so we did the same with our falling
factorial transforms to even the comparison.  For each problem size
$n$, we chose evenly spaced inputs (this is required for the
Fourier and wavelet transforms, but recall, not for the falling
factorial transform), and averaged the results over 10 repetitions.
The figure clearly
demonstrates a linear scaling for the runtimes of the falling
factorial transform, which matches their theoretical $O(n)$
complexity; the wavelet and fast fourier transforms also behave as
expected, with the former having $O(n)$ complexity, and the latter
$O(n\log{n})$.  In fact, a raw comparison of times shows that our
implementation of the falling factorial transforms runs slightly
faster than the highly-optimized wavelet transforms from the
Stanford WaveLab.

For completeness, Figure \ref{fig:runtimes2} displays a comparison
between the falling factorial transforms and the corresponding
transforms using the truncated power basis (also with $k=3$).  We see
that the latter scale quadratically with $n$, which is again to be
expected, as the truncated power basis matrix is essentially lower
triangular.

\subsection{Proximity to truncated power basis}
\label{sec:hgdiff}

With computational efficiency having been assured by the last lemma,
our next lemma lays the footing for the statistical credibility of the
falling factorial basis.

\begin{lemma}
\label{lem:hgdiff}
Let $G^{(k)}$ and $H^{(k)}$ be the $k$th order truncated power and
falling factorial matrices, defined
over inputs $0 \leq x_1 < \ldots < x_n \leq 1$.
Let $\delta = \max_{i=1,\ldots n} (x_i-x_{i-1})$, where we write
$x_0=0$. Then
\begin{equation*}
\max_{i,j=1,\ldots n} \, |G^{(k)}_{ij}-H^{(k)}_{ij}| \leq k^2\delta.
\end{equation*}
\end{lemma}

This tight elementwise bound between the two basis matrices will be
used in Section \ref{sec:tf} to prove a result on the convergence of
trend filtering estimates.  We will also discuss its importance in the
context of a fast nonparametric two-sample test in Section
\ref{sec:ks}.  To give a preview: in many problem instances, the
maximum gap $\delta$ between adjacent sorted inputs $x_1,\ldots x_n$
is of the order $\log{n}/n$ (for a more precise statement see Lemma
\ref{lem:randgap}), and this means that the maximum absolute
discrepancy between the elements of $G^{(k)}$ and $H^{(k)}$ decays
very quickly.

\section{Why not just use B-splines?}
\label{sec:bsplines}

B-splines already provide a computationally
efficient parametrization for the set of $k$th order splines; i.e.,
since they produce banded basis matrices, we can already perform
linear-time basis matrix multiplication and inversion with B-splines.
To confirm this point empirically, we included B-splines in the timing
comparison of Section \ref{sec:fastcomp}, refer to Figure
\ref{fig:runtimes1} for the results.
So, why not always use B-splines in place of the falling factorial
basis, which only approximately spans the space of splines?

A major reason is that the falling factorial functions (like the
truncated power functions) admit a sparse representation under the
total variation operator, whereas the B-spline functions do not.
To be more specific, suppose that $f_1,\ldots f_m$ are $k$th order
piecewise polynomial functions with knots at the points
$0 \leq z_1 < \ldots < z_r \leq 1$, where $m = r+k+1$.
Then, for $f=\sum_{j=1}^m \alpha_j f_j$, we have
\begin{equation*}
\TV(f^{(k)}) = \sum_{i=1}^r \left|\sum_{j=1}^m
\left(f_j^{(k)}(z_i) - f_j^{(k)}(z_{i-1})\right) \cdot \alpha_j \right|,
\end{equation*}
denoting $z_0=0$ for ease of notation.  If $f_1,\ldots f_m$
are the falling factorial functions defined over the points
$z_1,\ldots z_r$, then the term
\smash{$f_j^{(k)}(z_i)-f_j^{(k)}(z_{i-1})$} is equal to 0
for all $i,j$, except when $i=j-k-1$ and $j \geq k+2$, in
which case it equals 1.  Therefore,
\smash{$\TV(f^{(k)}) = \sum_{j=k+2}^m |\alpha_j|$}, a simple sum of
absolute coefficients in the falling factorial expansion.  The same
result holds for the truncated power basis functions.
But if $f_1,\ldots f_m$ are B-splines, then this is not true;
one can show that in this case
\smash{$\TV(f^{(k)}) = \|C\alpha\|_1$},
 where $C$ is a
(generically) dense matrix.
The fact that $C$ is dense makes it
cumbersome, both mathematically and computationally, to use the
B-spline parametrization in spline problems
involving total variation, such as those discussed in Sections
\ref{sec:tf} and \ref{sec:ks}.

\section{Trend filtering for arbitrary inputs}
\label{sec:tf}

Trend filtering is a relatively new method for nonparametric
regression. Suppose that we observe
\begin{equation}
\label{eq:regmodel}
y_i = f_0(x_i) + \epsilon_i,\;\;\; i=1,\ldots n,
\end{equation}
for a true (unknown) regression function $f_0$,
inputs $x_1 < \ldots < x_n \in \R$, and errors
$\epsilon_1,\ldots \epsilon_n$.  The trend filtering estimator was
first proposed by \citet{l1tf}, and further studied by
\citet{trendfilter}.
In fact, the latter work motivated the current paper, as it derived
properties of the falling factorial basis over evenly spaced
inputs $x_i=i/n$, $i=1,\ldots n$, and use these to prove convergence
rates for trend filtering estimators.  In the present section, we
allow $x_1,\ldots x_n$ to be arbitrary, and extend the convergence
guarantees for trend filtering, utilizing the properties of the
falling factorial basis derived in Section \ref{sec:props}.

\begin{figure}[tb]
\centering
\includegraphics[width=0.8\linewidth]{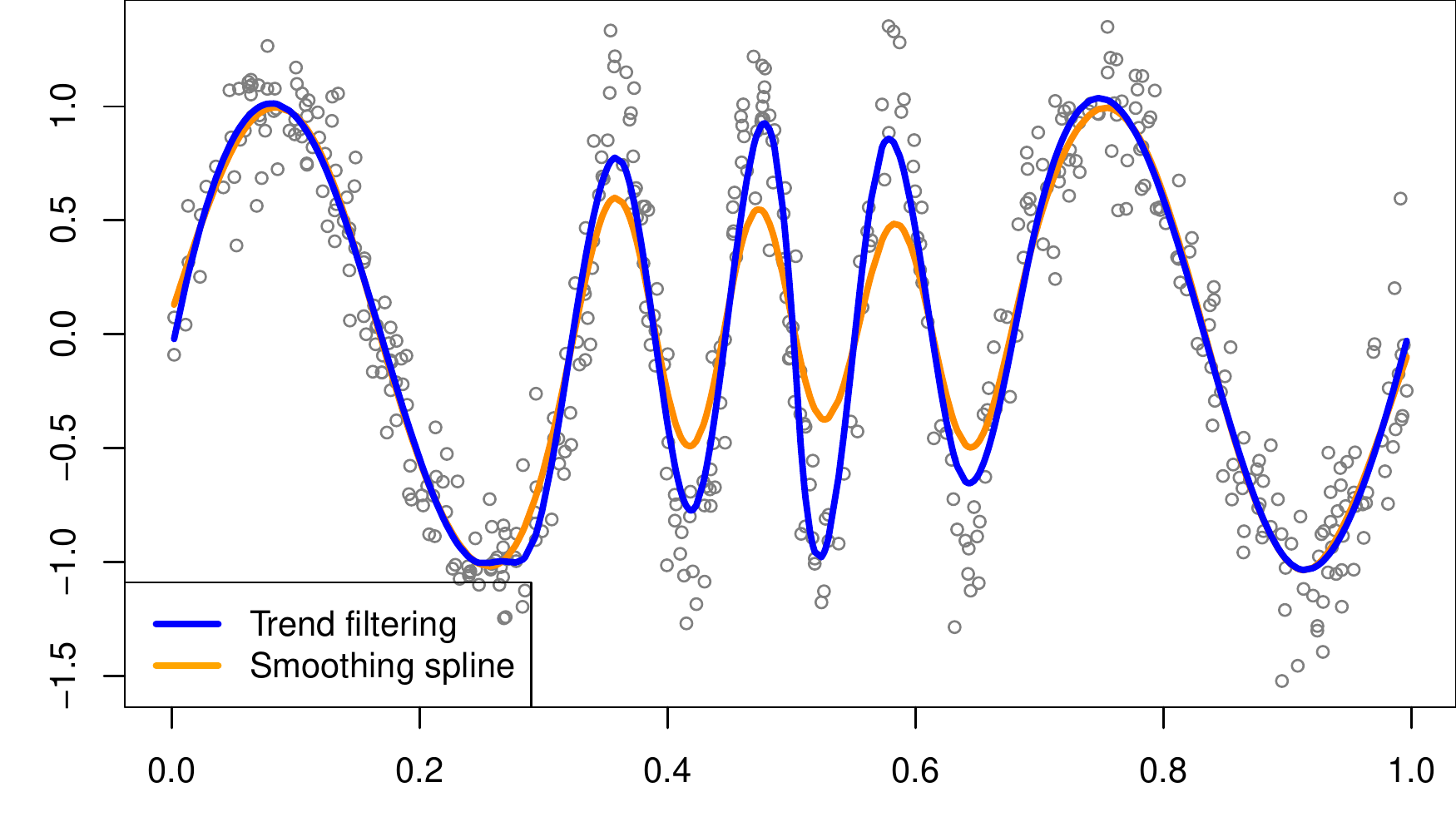}
\caption{Example trend filtering and smoothing spline
  estimates.}
\label{fig:tfex}
\end{figure}

The trend filtering estimate \smash{$\hbeta$} of order $k \geq 0$ is
defined by 
\begin{equation}
\label{eq:tf}
\hbeta = \argmin_{\beta \in \R^n} \,
\half\|y-\beta\|_2^2 + \lambda \cdot
\frac{1}{k!} \|D^{(k+1)} \beta\|_1,
\end{equation}
where $y=(y_1,\ldots y_n) \in \R^n$,
$D^{(k+1)} \in \R^{(n-k-1)\times n}$ is the $(k+1)$st order discrete
difference operator defined in \eqref{eq:dk} over the
input points $x_1,\ldots x_n$, and $\lambda \geq 0$ is a tuning
parameter.  We can think of the components of
\smash{$\hbeta$} as defining an estimated function
\smash{$\hf$} over the input points.
To give an example, in Figure \ref{fig:tfex}, we drew noisy
observations from a smooth underlying function, where the input
points $x_1,\ldots x_n$ were sampled
uniformly at random over $[0,1]$, and we computed the trend filtering
estimate \smash{$\hbeta$} with $k=3$ and a particular choice of
$\lambda$. From the plot (where we interpolated between
\smash{$(x_1,\hbeta_1),\ldots (x_n,\hbeta_n)$}
for visualization purposes), we can see that the implicitly defined
trend filtering function \smash{$\hf$} displays a piecewise cubic
structure, with adaptively chosen knot points.
Lemma \ref{lem:hinv} makes this connection precise by showing that
such a function $\hf$ is indeed a linear combination of falling
factorial functions.  Letting $\beta=H^{(k)}\alpha$, where $H^{(k)}
\in \R^{n\times n}$ is the $k$th order falling factorial basis matrix
defined over the inputs $x_1,\ldots x_n$, the trend filtering problem
in \eqref{eq:tf} becomes
\begin{equation}
\label{eq:tfh}
\halpha = \argmin_{\alpha \in \R^n} \,
\half \|y-H^{(k)} \alpha\|_2^2 +
\lambda \cdot \hspace{-3pt} \sum_{j=k+2}^n |\alpha_j|,
\end{equation}
equivalent to the functional minimization problem
\begin{equation}
\label{eq:tfcont}
\hf = \argmin_{f \in \cH_k} \,
\half \sum_{i=1}^n \big(y_i- f(x_i)\big)^2
+ \lambda \cdot \TV\big(f^{(k)}),
\end{equation}
where $\cH_k = \mathrm{span}\{h_1,\ldots h_n\}$ is the span of the
$k$th order falling factorial functions in \eqref{eq:hbasis},
$\TV(\cdot)$ denotes the total variation operator, and $f^{(k)}$
denotes the $k$th weak derivative of $f$.  In other words, the
solutions of problems \eqref{eq:tf} and \eqref{eq:tfcont} are related
by \smash{$ \hbeta_i=\hf(x_i)$}, $i=1,\ldots n$.  The trend
filtering estimate hence verifiably exhibits the structure of a
$k$th order piecewise polynomial function, with knots at
a subset of $x_1,\ldots x_n$, and this function is not necessarily
a spline, but is close to one (since it lies in the span of the
falling factorial functions $h_1,\ldots h_n$).

In Figure \ref{fig:tfex}, we also fit a smoothing spline estimate to
the same example data.  A striking difference: the trend filtering
estimate is far more locally adaptive towards the middle of
plot, where the underlying
function is less smooth (the two estimates
were tuned to have the same degrees of freedom, to even the
comparison).  This phenomenon is
investigated in \citet{trendfilter}, where it is shown that trend
filtering estimates attain the minimax convergence rate over a
large class of underlying functions, a class for which it is
known that smoothing splines (along with any other estimator linear in
$y$) are  suboptimal.  This latter work focused on evenly
spaced inputs, $x_i=i/n$, $i=1,\ldots n$, and the next two subsections
extend the trend filtering convergence theory to cover arbitrary
inputs $x_1,\ldots x_n \in [0,1]$.  We first consider the input points
as fixed, and then random.
All proofs are deferred until the appendix.

\subsection{Fixed input points}

The following is our main result on trend filtering.

\begin{theorem}
\label{thm:fixedconv}
Let $y\in\R^n$ be drawn from \eqref{eq:regmodel},
with fixed inputs $0 \leq x_1<\ldots <x_n \leq 1$, having
a maximum gap
\begin{equation}
\label{eq:maxgap}
\max_{i=1,\ldots n} \, (x_i-x_{i-1}) = O(\log{n}/n),
\end{equation}
and i.i.d., mean zero sub-Gaussian errors. Assume
that, for an integer $k \geq 0$ and constant $C>0$, the true
function $f_0$ is $k$ times weakly differentiable, with
\smash{$\TV(f_0^{(k)}) \leq C$}.
Then the $k$th order trend filtering estimate \smash{$\hbeta$} in
\eqref{eq:tf}, with tuning parameter value
$\lambda=\Theta(n^{1/(2k+3)})$, satisfies
\begin{equation}
\label{eq:conv}
\frac{1}{n}\sum_{i=1}^n \big(\hbeta_i-f_0(x_i)\big)^2 =
O_\P(n^{-(2k+2)/(2k+3)}).
\end{equation}
\end{theorem}

\noindent
{\it Remark 1.}  The rate \smash{$n^{-(2k+2)/(2k+3)}$} is the minimax
rate of convergence with respect to the class of $k$ times weakly
differentiable functions $f$ such that $\TV(f^{(k)}) \leq C$ (see,
e.g., \citet{nussbaum}, \citet{trendfilter}).  Hence
Theorem \ref{thm:fixedconv} shows that trend filtering estimates
converge at the minimax rate over a broad class of true functions
$f_0$, assuming that the fixed input points are not too irregular, in
that the maximum adjacent gap between points must satisfy
\eqref{eq:maxgap}.  This condition is not stringent and is naturally
satisfied by continuously distributed random inputs, as we show in the
next subsection.  We note that \citet{trendfilter} proved the same
conclusion (as in Theorem \ref{thm:fixedconv}) for unevenly spaced
inputs $x_1,\ldots x_n$, but placed very complicated and basically
uninterpretable conditions on the inputs.  Our
tighter analysis of the falling factorial functions yields the
simple sufficient condition \eqref{eq:maxgap}.

\smallskip
\smallskip
\noindent
{\it Remark 2.}  The conclusion in the theorem can be
strengthened, beyond the the convergence of \smash{$\hbeta$} to $f_0$
in \eqref{eq:conv}; under the same assumptions, the trend
filtering estimate \smash{$\hbeta$} also converges to
\smash{$\hf^\mathrm{spline}$} at the same rate
\smash{$n^{-(2k+2)/(2k+3)}$},
where we write \smash{$\hf^\mathrm{spline}$} to denote the solution in
\eqref{eq:tfcont} with $\cH_k$ replaced by
\smash{$\cG_k=\mathrm{span}\{g_1,\ldots g_n\}$}, the span of the
truncated power basis functions in \eqref{eq:gbasis}. This asserts
that the trend filtering estimate is indeed ``close to'' a spline, and
here the bound in Lemma \ref{lem:hgdiff}, between the truncated
power and falling factorial basis matrices, is key.
Moreover, we actually rely on the convergence of \smash{$\hbeta$} to
\smash{$\hf^\mathrm{spline}$} to establish
\eqref{eq:conv}, as the total variation regularized spline
estimator \smash{$\hf^\mathrm{spline}$} is already known to converge
to $f_0$ at the minimax rate \citep{locadapt}.


\subsection{Random input points}

To analyze trend filtering for random inputs, $x_1,\ldots x_n$,
we need to bound the maximum gap between adjacent points with high
probability.  Fortunately, this is possible for a
large class of distributions, as shown in the next lemma.

\begin{lemma}
\label{lem:randgap}
If $0 \leq x_1 < \ldots < x_n \leq 1$ are sorted i.i.d.\ draws from
an arbitrary continuous distribution supported on $[0,1]$, whose
density is bounded below by $p_0 > 0$, then with probability at
least $1-2p_0n^{-10}$,
\begin{equation*}
\max_{i=1,\ldots n} \, (x_i-x_{i-1}) \leq \frac{c_0\log n}{p_0n},
\end{equation*}
for a universal constant $c_0$.
\end{lemma}
The proof of this result is readily assembled from classical results
on order statistics; we give a simple alternate proof
in the appendix.
Lemma \ref{lem:randgap} implies the next corollary.


\begin{corollary}
\label{cor:randomconv}
Let $y\in\R^n$ be distributed according to the model
\eqref{eq:regmodel}, where the inputs $0 \leq x_1 < \ldots < x_n \leq
1$ are sorted i.i.d.\ draws from an arbitrary continuous distribution
on $[0,1]$, whose density is bounded below.  Assume again that the
errors are i.i.d., mean zero sub-Gaussian variates, independent of the
inputs, and that the true function $f_0$ has $k$ weak derivatives
and satisfies \smash{$\TV(f_0^{(k)}) \leq C$}.  Then, for
$\lambda=\Theta(n^{1/(2k+3)})$, the $k$th order
trend filtering estimate \smash{$\hbeta$} converges at the same
rate as in Theorem \ref{thm:fixedconv}.
\end{corollary}

\section{A higher order Kolmogorov-Smirnov test}
\label{sec:ks}

The two-sample Kolmogorov-Smirnov (KS) test is a standard
nonparametric hypothesis test of equality between two distributions,
say $\P_X$ and $\P_Y$, from independent samples $x_1,\ldots x_m
\sim \P_X$ and $y_1,\ldots y_n \sim \P_Y$.  Writing
$X_{(m)}=(x_1,\ldots x_m)$, 
$Y_{(n)}=(y_1,\ldots y_n)$, 
and $Z_{(m+n)}=(z_1,\ldots z_{m+n})=X_{(m)} \cup Y_{(n)}$ for the
joined samples, the KS statistic can be expressed as
\begin{equation}
\label{eq:ks}
\KS(X_{(m)},Y_{(n)})= \max_{z_j \in Z_{(m+n)}} \,
\left|\frac{1}{m}\sum_{i=1}^m 1\{x_i \leq z_j\} -
\frac{1}{n}\sum_{i=1}^n 1\{y_i \leq z_j\} \right|.
\end{equation}
This examines the maximum absolute difference between the
empirical cumulative distribution functions from $X_{(m)}$ and
$Y_{(n)}$, across all points in the joint set $Z_{(m+n)}$,
and so the test rejects for large values of \eqref{eq:ks}.
A well-known alternative (variational) form for the KS statistic is
\begin{equation}
\label{eq:ksvar}
\KS(X_{(m)},Y_{(n)}) = \max_{f \,:\, \TV(f) \leq 1} \,
\left|\hat{\E}_{X_{(m)}}[f(X)] - \hat{\E}_{Y_{(n)}} [f(Y)]\right|,
\end{equation}
where \smash{$\hat{\E}_{X_{(m)}}$} denotes the empirical expectation
under $X_{(m)}$, so that
\smash{$\hat{\E}_{X_{(m)}} [f(X)] = 1/m \sum_{i=1}^m f(x_i)$}, and
similarly for \smash{$\hat{\E}_{Y_{(n)}}$}.  The equivalence between
\eqref{eq:ksvar} and \eqref{eq:ks} comes from the fact that maximum in
\eqref{eq:ksvar} is achieved by taking $f$ to be a step function, with
its knot (breakpoint) at one of the joined samples $z_1,\ldots z_{m+n}$.

The KS test is perhaps one of the most widely used nonparametric tests
of distributions, but it does have its shortcomings.  Loosely speaking,
it is known to be sensitive in detecting differences between the
centers of distributions $\P_X$ and $\P_Y$, but much less sensitive in
detecting differences in the tails.  In this section, we generalize
the KS test to ``higher order'' variants that are more
powerful than the original KS test in detecting tail differences
(when, of course, such differences are present).  We first define
the higher order KS test, and describe how it can be computed
in linear time with the falling factorial basis.  We then empirically
compare these higher order versions to the original KS test, and
several other commonly used nonparametric two-sample tests of
distributions.



\subsection{Definition of the higher order KS tests}
\label{sec:higherorderKS}


For a given order $k \geq 0$, we define the $k$th order KS test
statistic between $X_{(m)}$ and $Y_{(n)}$ as
\begin{equation}
\label{eq:kskg}
\KS_G^{(k)}(X_{(m)},Y_{(n)}) = \left\| (G_2^{(k)})^T \left(
\frac{ \mathds{1}_{X_{(m)}} }{m} -
\frac{ \mathds{1}_{Y_{(n)}} }{n} \right) \right\|_\infty.
\end{equation}
Here \smash{$G^{(k)} \in \R^{(m+n)\times (m+n)}$} is the
$k$th order
truncated power basis matrix over the joined samples
$z_1 < \ldots < z_{m+n}$, assumed sorted without a
loss of generality, and
\smash{$G_2^{(k)}$} is the submatrix 
formed by excluding its first $k+1$ columns.  Also,
\smash{$\mathds{1}_{X_{(m)}} \in \R^{(m+n)}$} is a vector whose
components indicate the locations of $x_1<\ldots <x_m$ among
$z_1<\ldots< z_{m+n}$, and similarly for
\smash{$\mathds{1}_{Y_{(n)}}$}.  Finally, $\|\cdot\|_\infty$ denotes
the $\ell_\infty$ norm, $\|u\|_\infty = \max_{i=i,\ldots r} |u_i|$ for
$u \in \R^r$.

As per the spirit of our paper, an alternate definition for the
$k$th order KS statistic uses the falling factorial basis,
\begin{equation}
\label{eq:kskh}
\KS_H^{(k)}(X_{(m)},Y_{(n)}) = \left\| (H_2^{(k)})^T \left(
\frac{ \mathds{1}_{X_{(m)}} }{m} -
\frac{ \mathds{1}_{Y_{(n)}} }{n}\right) \right\|_\infty,
\end{equation}
where now \smash{$H^{(k)} \in \R^{(m+n)\times (m+n)}$} is the
$k$th order falling factorial basis matrix over the joined samples
$z_1 < \ldots < z_{m+n}$.  Not surprisingly, the two definitions are
very close, and H\"{o}lder's inequality shows that
\begin{equation*}
|\KS_G^{(k)}(X_{(m)},Y_{(n)}) - \KS_H^{(k)}(X_{(m)},Y_{(n)})|
\leq \max_{i,j=1,\ldots m+n}\, 2|G^{(k)}_{ij} - H^{(k)}_{ij} | \leq
2k^2 \delta,
\end{equation*}
the last inequality due to Lemma \ref{lem:hgdiff}, with $\delta$
the maximum gap between $z_1,\ldots z_{m+n}$.  Recall that
Lemma \ref{lem:randgap} shows $\delta$ to be of the order
$\log(m+n)/(m+n)$
for continuous distributions $\P_X,\P_Y$ supported nontrivially on
$[0,1]$, which means that with high probability, the two definitions
differ by at most $2k^2 \log(m+n)/(m+n)$, in such a setup.

The advantage to using the falling factorial definition is that
the test statistic in \eqref{eq:kskh} can be computed in $O(k(m+n))$
time, without even
having to form the matrix \smash{$H^{(k)}_2$}
(this is assuming sorted points $z_1,\ldots z_{m+n}$).  See Lemma
\ref{lem:fastcomp}, and Algorithm \ref{alg:hkt} in the appendix.  By
comparison, the statistic in \eqref{eq:kskg} requires $O((m+n)^2)$
operations.  In addition to the theoretical bound described above, we
also find empirically that the two definitions perform quite
similarly, as shown in the next subsection, and hence we advocate the
use of \smash{$\KS^{(k)}_H$} for computational reasons.

A motivation for our proposed tests is as follows: it can be shown
that \eqref{eq:kskg}, and therefore \eqref{eq:kskh}, approximately
take a variational form similar to \eqref{eq:ksvar}, but where the
constraint is over functions whose $k$th (weak) derivative has total
variation at most 1.  See the appendix.

\subsection{Numerical experiments}

\begin{figure}[tb]
  \centering
  \includegraphics[width=0.48\linewidth]{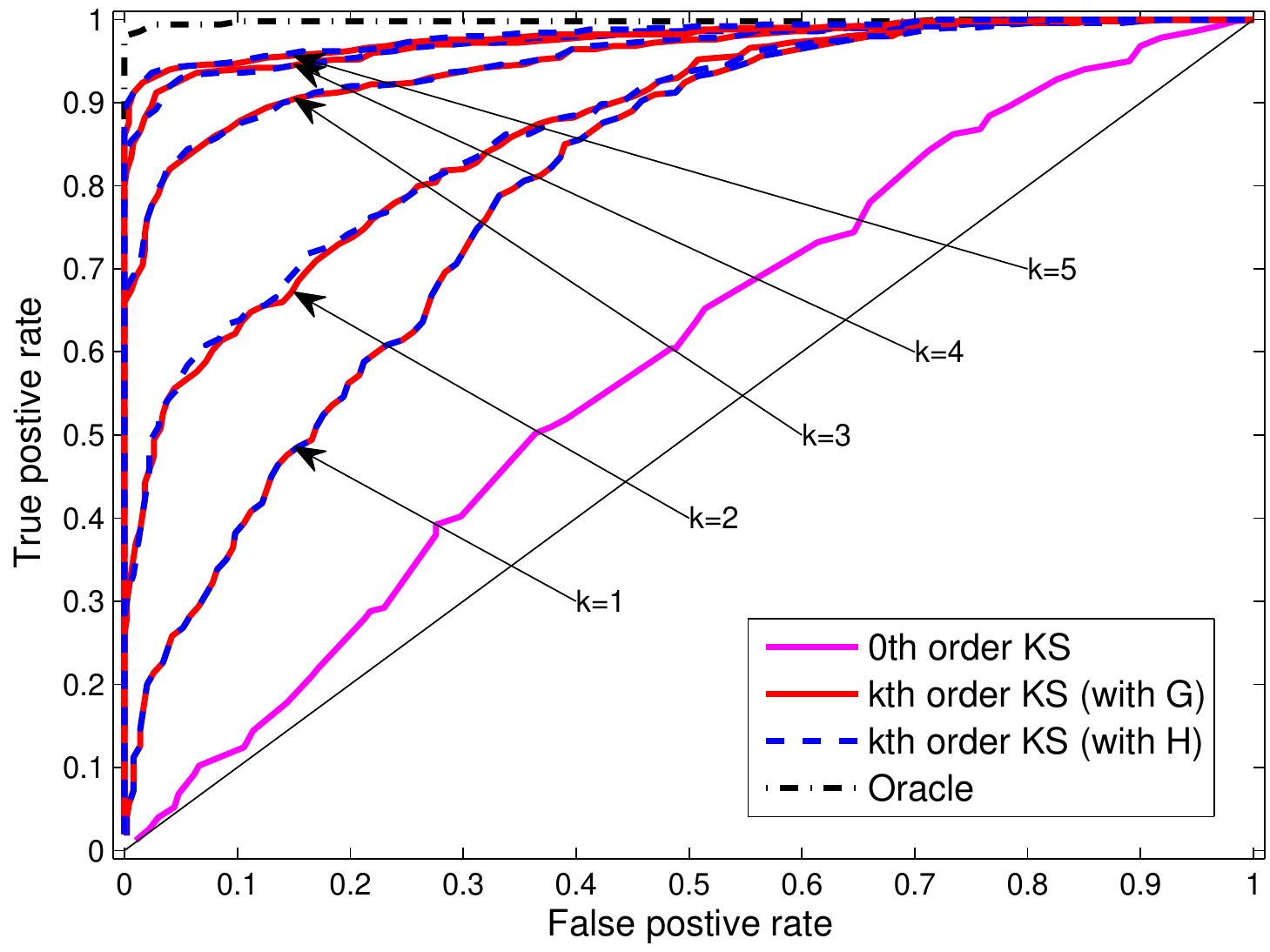}
  \includegraphics[width=0.48\linewidth]{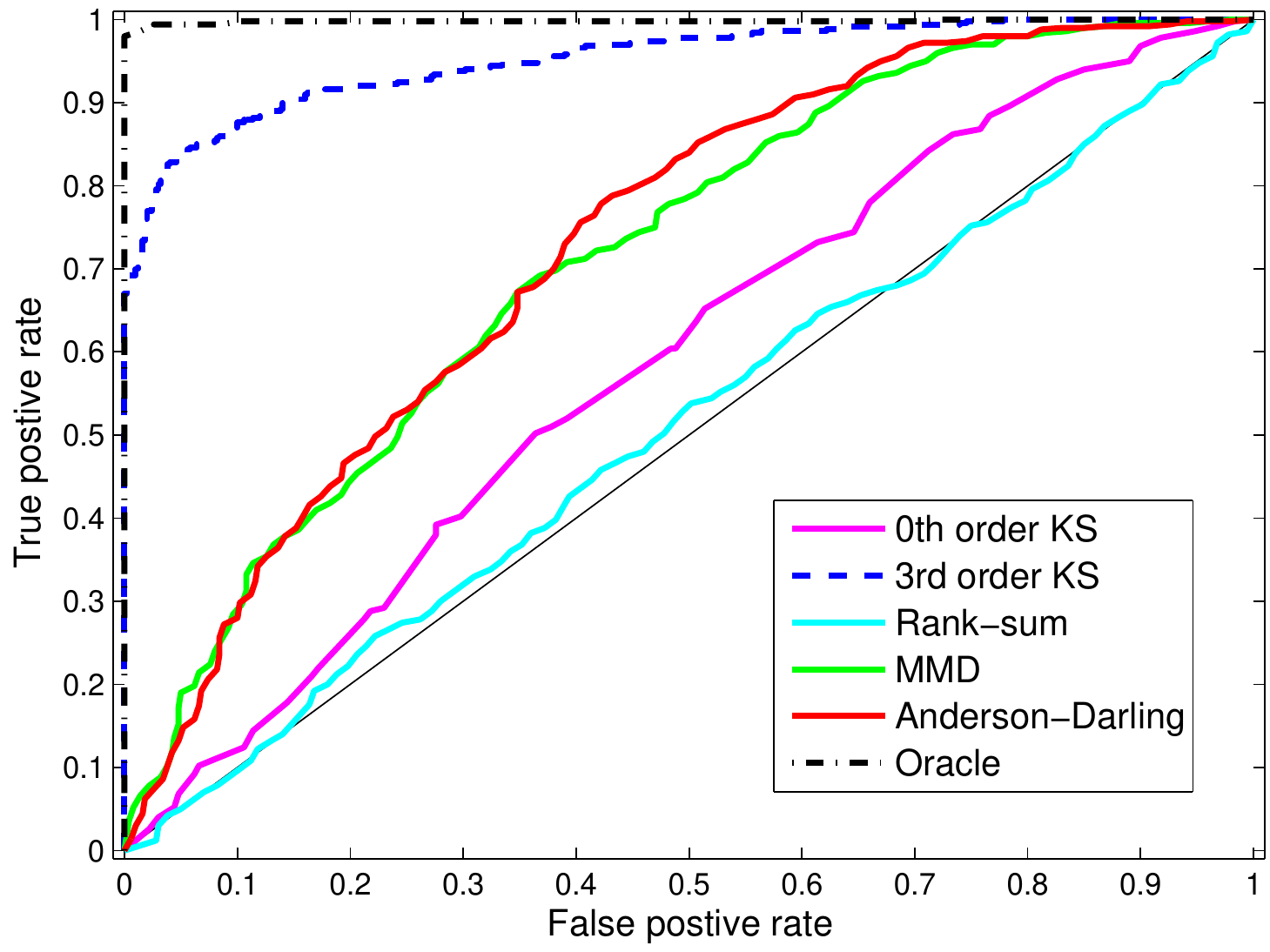}\\
    \vspace{-10pt}
  \caption{ROC curves for experiment 1, normal vs.\ t.}
\label{fig:exp1}
\end{figure}

\begin{figure}[tb]
  \centering
  \includegraphics[width=0.48\linewidth]{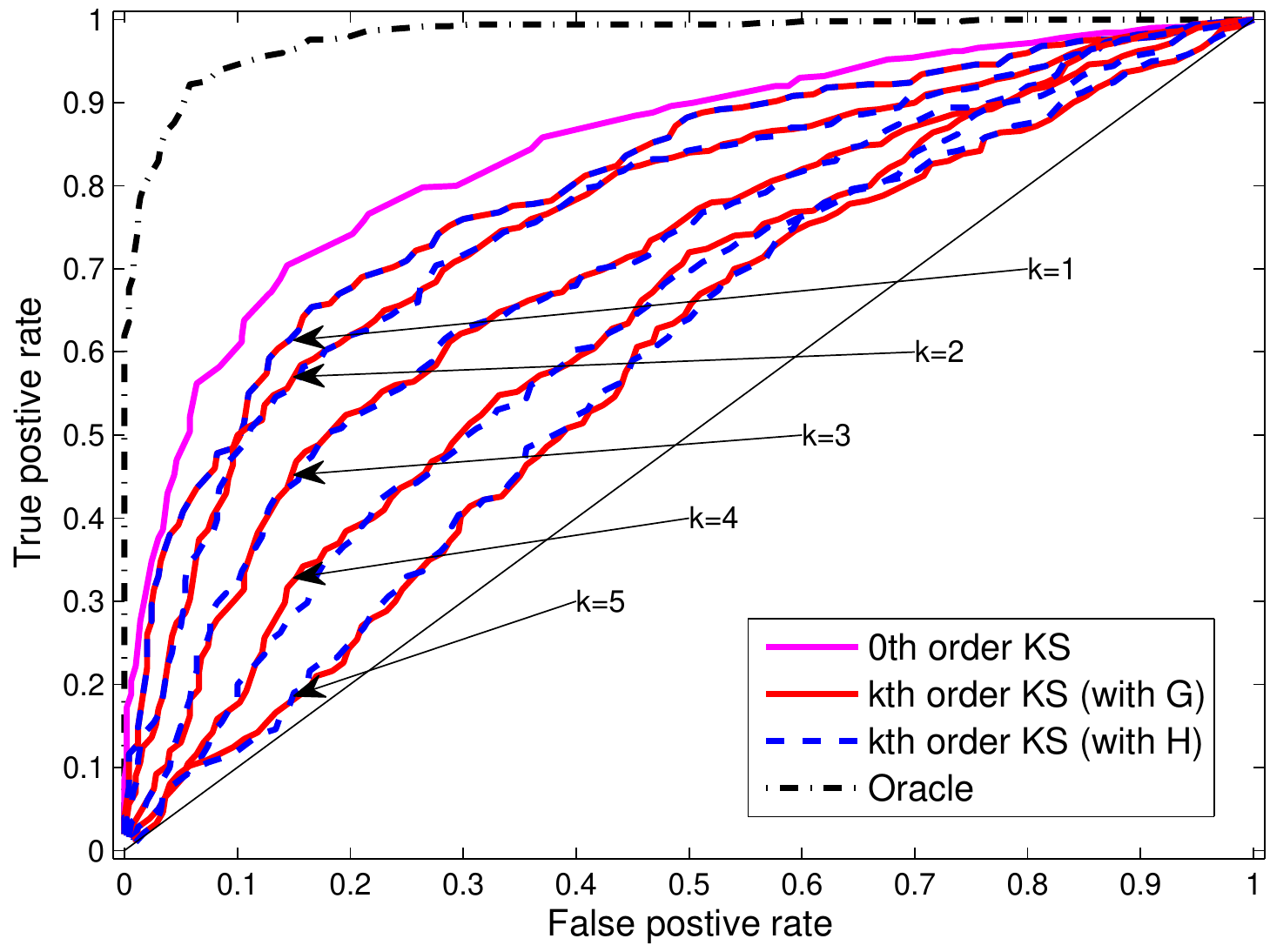}
  \includegraphics[width=0.48\linewidth]{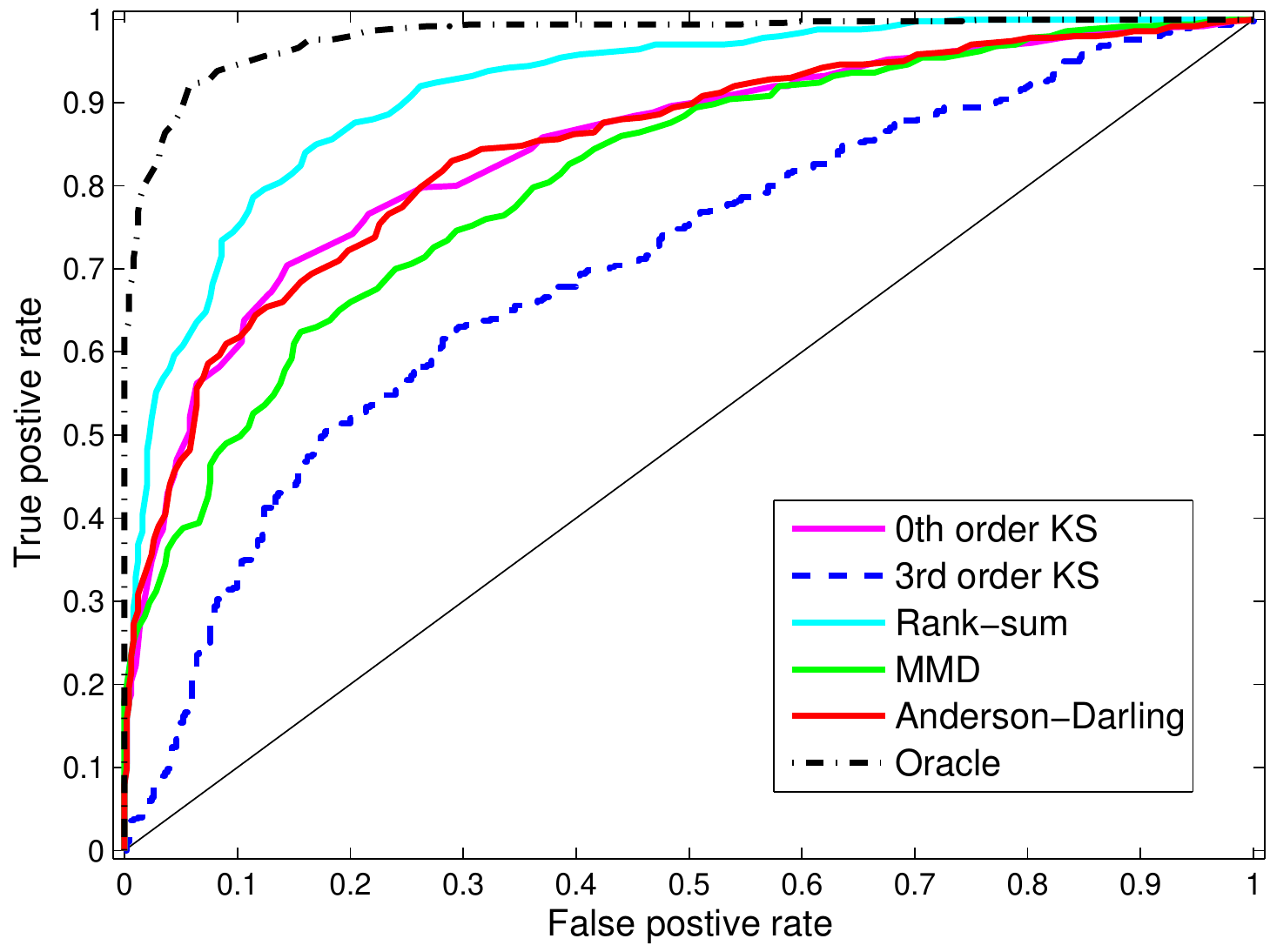}\\
  \vspace{-10pt}
  \caption{ROC curves for experiment 2, Laplace vs.\ Laplace.}
  \label{fig:exp2}
\end{figure}


We examine the higher order KS tests
by simulation.
The setup: we fix two distributions $P,Q$.  We
draw $n$ i.i.d.\ samples $X_{(n)},Y_{(n)} \sim P$,
calculate a test statistic, and repeat this $R/2$ times;
we also draw $n$ i.i.d.\ samples $X_{(n)} \sim P$, $Y_{(n)} \sim
Q$, calculate a test statistic, and repeat $R/2$ times.
We then construct an ROC curve, i.e.,
the true positive rate versus the
false positive rate of the test, as we vary its rejection threshold.
For the test itself, we consider our $k$th order KS test, in
both its $G$ and $H$ forms, as well as the usual KS test, and a number
of other popular two-sample tests: the Anderson-Darling test
\citep{anderson1954test,scholz1987k}, the Wilcoxon rank-sum test
\citep{wilcoxon1945individual}, and the maximum mean
discrepancy (MMD) test, with RBF kernel \citep{gretton2012kernel}.


Figures \ref{fig:exp1} and \ref{fig:exp2} show
the results of two experiments in which $n=100$
and $R=1000$. (See the appendix for more experiments.)
In the first we used $P=N(0,1)$ and $Q=t_3$ ($t$-distribution with 3
degrees of freedom), and in the second $P=\mathrm{Laplace}(0)$ and
$Q=\mathrm{Laplace}(0.3)$ (Laplace distributions of different
means).
We see that our proposed $k$th order KS test performs favorably in the
first experiment, with its power increasing with $k$.  When $k=3$, it
handily beats all competitors in detecting the difference between the
standard normal distribution and the heavier-tailed $t$-distribution.
But there is no free lunch: in the second experiment, where the
differences between $P,Q$ are mostly near the centers of the
distributions and not  in the tails, we can see that increasing $k$
only decreases the power of the $k$th order KS test.
 In short, one can view our proposal as
introducing a family of tests parametrized by $k$, which offer a
tradeoff in center versus tail sensitivity. A more thorough study
will be left to future work.

\section{Discussion}
\label{sec:discuss}
We formally proposed and analyzed the spline-like
falling factorial basis functions.  These basis functions admit
attractive computational and statistical properties, and
we demonstrated their applicability in two problems:
trend filtering, and a novel higher order variant of the KS test.
These examples, we feel, are just the beginning. As typical
operations associated with the falling factorial basis scale merely
linearly with the input size (after sorting), we feel that this basis
may be particularly well-suited to a rich number of large-scale
applications in the modern data era, a direction that we are excited
to pursue in the future.



\smallskip\smallskip
\noindent
{\bf Acknowledgements}
The research was partially supported by NSF Grant DMS-1309174, Google Faculty Research Grant and the Singapore National Research Foundation under its International Research Centre @ Singapore Funding Initiative and administered by the IDM Programme Office.

\appendix
\appendixpage
\addappheadtotoc


This appendix contains proofs and additional experiments for the paper ``The Falling Factorial Basis and Its Statistical Applications''.
In Section~\ref{app:proofs}, we provide
proofs to the key technical results in the main paper.  In
Section~\ref{app:ks}, we give some motivating arguments and
additional experiments for the higher order KS test.

\section{Proofs and technical details}
\label{app:proofs}

\subsection{Proof of Lemma \ref{lem:recursive} (recursive
  decomposition)}
\label{A.sec:cumsum}

The falling factorial basis matrix, as defined in \eqref{eq:hbasis},
\eqref{eq:h}, can be expressed as
\smash{$H^{(k)}=[H^{(k)}_1 \, H^{(k)}_2]$}, where
\begin{equation*}
  H^{(k)}_1 = \left[\begin{array}{ccccc}
    1     & 0           & 0                           & \cdots     & 0 \\
    1     & x_2-x_1     & 0                           & \cdots     & 0 \\
    1     & x_3-x_1     & (x_3-x_2)(x_3-x_1)& \cdots     & \vdots \\
    \vdots& \vdots      & \vdots                      & \ddots    & \vdots \\
    1     & x_{k+1}-x_1      & (x_{k+1}-x_2) (x_{k+1}-x_1)   & \cdots    & \prod_{\ell=1}^{k}(x_{k+1}-x_{\ell}) \\
    \vdots& \vdots      & \vdots                      & \ddots    & \vdots \\
    1     & x_n-x_1 & (x_n-x_2) (x_n-x_1)   & \cdots     & \prod_{\ell=1}^{k} (x_n-x_{\ell}) \\
  \end{array}\right] \in \R^{n \times (k+1)},
\end{equation*}
and
\begin{equation*}
  H^{(k)}_2 =  \left[\begin{array}{cccc}
    0_{(k+1)\times 1}     & 0_{(k+1)\times 1}   & \cdots  & 0_{(k+1)\times 1}   \\
    \prod_{\ell=1}^{k} (x_{k+2}-x_{1+\ell})  & 0          & \cdots     & 0 \\
    \prod_{\ell=1}^{k} (x_{k+3}-x_{1+\ell})  & \prod_{\ell=1}^{k} (x_{k+3}-x_{2+\ell}) &  \cdots     & 0 \\
    \vdots& \vdots                            & \ddots    & \vdots \\
    \prod_{\ell=1}^{k} (x_{n}-x_{1+\ell})     & \prod_{\ell=1}^{k} (x_{n}-x_{2+\ell})        &\cdots & \prod_{\ell=1}^{k} (x_n-x_{n-k-1+\ell}) \\
  \end{array}\right]\in \R^{n \times (n-k-1)}.
\end{equation*}
Lemma~\ref{lem:recursive} claims
that \smash{$H^{(0)}=L_n$}, the lower triangular matrix of $1$s, which
can be seen directly by inspection (recalling our convention of
defining thee empty product to be 1).  The lemma further claims that
\smash{$H^{(k)}$} can be recursively factorized into the following
form:
\begin{equation}
\label{A.eq:hk}
H^{(k)} = H^{(k-1)} \cdot
\left[\begin{array}{cc}
I_k & 0 \\ 0 & \Delta^{(k)}
\end{array}\right]
\cdot
\left[\begin{array}{cc}
I_k & 0 \\ 0 & L_{n-k}
\end{array}\right],
\end{equation}
for all $k\geq 1$.  We prove the above factorization in this current
section. In what follows,
we denote the last $n-k-1$ columns of the product
\eqref{A.eq:hk} by \smash{$\tilde{M}^{(k)} \in \R^{n\times(n-k-1)}$},
and also write
$$
\tilde{M}^{(k)} = \left[\begin{array}{c}
0_{(k+1)\times (n-k-1)} \smallskip\\
\tilde{L}^{(k)},
\end{array}\right],
$$
i.e., we use \smash{$\tilde{L}^{(k)}$} to denote the lower $(n-k-1)
\times (n-k-1)$ submatrix of \smash{$\tilde{M}^{(k)}$}.  To prove the
lemma, we show that \smash{$\tilde{M}^{(k)}$} is equal to
the corresponding block \smash{$H^{(k)}_2$}, by
induction on $k$.  The proof that the first block of $k+1$ columns of
the product is equal to \smash{$H^{(k)}_1$} follows from the arguments
given for the proof of the second block, and therefore we do not
explicitly rewrite the proof for this part.

We begin the inductive proof by checking the case $k=1$.  Note
\begin{align}
\nonumber
\tilde{M}^{(1)} =
\left[\begin{array}{c}
  0_{2\times (n-2)} \smallskip \\
  \tilde{L}^{(1)}
\end{array}\right] &=
\left[\begin{array}{c}
    0_{1\times (n-1)}\\
    L_{n-1}
\end{array}\right]
  (\Delta^{(k)})^{-1}
\left[\begin{array}{c}
    0_{1\times (n-2)}\\
    L_{n-2} \\
\end{array}\right] \\
\nonumber &=
\left[\begin{array}{cccc}
    0_{2\times 1}
    &0_{2\times 1} & \cdots  &0_{2\times 1}\\
    x_3-x_2& 0 & \cdots & 0 \\
    x_4-x_2& x_4-x_3 & \cdots &0 \\
    \vdots& \vdots& \ddots& \vdots \\
    x_n-x_2& x_n-x_3&\cdots&x_n-x_{n-1}
\end{array}\right].
\end{align}
This gives precisely the last $n-2$ columns of $H^{(1)}$, as defined
in \eqref{eq:hbasis}.


Next we verify that if the statement holds for some $k \geq 1$, then it is true for $k+1$.
To avoid confusion, we will use $i,j$ as indices
\smash{$H^{(k+1)}$} and $\alpha,\beta$ as indices of
\smash{$\tilde{L}^{(k+1)}$}. The universal rule for the relationship
between the two sets of indices is
$$ \begin{pmatrix}
     i \\
     j \\
   \end{pmatrix}=\begin{pmatrix}
                   \alpha \\
                   \beta \\
                 \end{pmatrix}+k+2.
$$
We consider an arbitrary element,
\smash{$\tilde{L}^{(k+1)}_{\alpha \beta}$}. Due to the upper triangular
  shape of $\tilde{L}^{(k)}$, we have
  $\tilde{L}^{(k)}_{\alpha\beta}=0$ if $\alpha<\beta$.  For $\alpha
  \geq \beta$, we plainly calculate, using the inductive hypothesis
\begin{align}
  \tilde{L}^{(k+1)}_{\alpha\beta}
&=\sum_{q=1+\beta}^{1+\alpha}\tilde{L}^{(k)}_{1+\alpha,q}\cdot
  (\Delta^{(k+1)})^{-1}_{qq} \nonumber \\
  &=  \sum_{q=1+\beta}^{1+\alpha}
    \prod_{\ell=1}^k(x_{k+2+\alpha}-x_{q+\ell})
\cdot (x_{k+1+q}-x_q) \nonumber \\
   &= \prod_{\ell=1}^{k+1}(x_{k+2+\alpha}-x_{\beta+\ell})\cdot
   A=H^{(k)}_{ij}\cdot A, \nonumber 
\end{align}
where $A$ is the sum of terms that scales each summand to the desired
quantity (by multiplying and dividing by missing factors). To complete
the inductive proof, it suffices to show that $A=1$. It turns out that
there are two main cases to consider, which we examine below.

{\it Case 1.}
When $\alpha-\beta\leq k$, the term $A$ can be expressed as
\begin{align*}
A=&\frac{x_{k+1+1+\beta}-x_{1+\beta}}{x_{k+2+\alpha}-x_{1+\beta}}+\frac{(x_{k+1+2+\beta}
-x_{2+\beta})(x_{k+2+\alpha}-x_{k+1+1+\beta})}{(x_{k+2+\alpha}-x_{1+\beta})(x_{k+2+\alpha}-x_{2+\beta})}\\
&+\cdots+\frac{(x_{k+1+\gamma+\beta}
-x_{\gamma+\beta})(x_{k+2+\alpha}-x_{k+2+\beta})\cdots(x_{k+2+\alpha}-x_{k+\gamma+\beta})}
{(x_{k+2+\alpha}-x_{1+\beta})\cdots(x_{k+2+\alpha}-x_{\gamma-1+\beta})(x_{k+2+\alpha}-x_{\gamma+\beta})}\\
&+\cdots+\frac{(x_{k+1+\alpha}
-x_{\alpha})(x_{k+2+\alpha}-x_{k+2+\beta})\cdots(x_{k+2+\alpha}-x_{k+1+\alpha})}
{(x_{k+2+\alpha}-x_{1+\beta})\cdots(x_{k+2+\alpha}-x_{\alpha-1})(x_{k+2+\alpha}-x_{\alpha})}\\
&+\frac{\cancel{(x_{k+2+\alpha}
-x_{1+\alpha})}(x_{k+2+\alpha}-x_{k+2+\beta})\cdots(x_{k+2+\alpha}-x_{k+1+\alpha})}
{(x_{k+2+\alpha}-x_{1+\beta})\cdots(x_{k+2+\alpha}-x_{\alpha})\cancel{(x_{k+2+\alpha}-x_{1+\alpha})}}.
\end{align*}
Note that in the last term, the factor $(x_{k+2+\alpha}-x_{1+\alpha})$
in both the denominator and numerator cancels out, leaving the
denominator to be the same as the second to last term.
Combining the last two terms, we again get a common factor
$(x_{k+2+\alpha}-x_{\alpha})$ in denominator and numerator, which cancels
out, and makes the denominator of this term the same as that previous
term.  Continuing in this manner, we can recursively eliminate the
terms from last to the first, leaving
$$ \frac{ \cancel{x_{k+2+\beta}}
  -x_{1+\beta}+x_{k+2+\alpha}-\cancel{x_{k+2+\beta}}
}{x_{k+2+\alpha}-x_{1+\beta}}=1.$$
In other words, we have shown that $A=1$.

{\it Case 2.} When $\alpha-\beta \geq k+1$, the denominators in terms
of $A$ will remain the same after they reach
$$(x_{k+2+\alpha}-x_{1+\beta})\cdots(x_{k+2+\alpha}-x_{1+k+\beta})
=\prod_{\ell=1}^{k+1}(x_{k+2+\alpha}-x_{\beta+\ell}):=B.$$
Again, we begin by expressing $A$ explicitly as
\begin{align}
A=&\frac{x_{k+1+1+\beta}-x_{1+\beta}}{x_{k+2+\alpha}-x_{1+\beta}}+\frac{(x_{k+1+2+\beta}
-x_{2+\beta})(x_{k+2+\alpha}-x_{k+1+1+\beta})}{(x_{k+2+\alpha}-x_{1+\beta})(x_{k+2+\alpha}-x_{2+\beta})}\nonumber\\
&+\cdots+\frac{(x_{k+1+\gamma+\beta}
-x_{\gamma+\beta})(x_{k+2+\alpha}-x_{k+2+\beta})\cdots(x_{k+2+\alpha}-x_{k+\gamma+\beta})}
{(x_{k+2+\alpha}-x_{1+\beta})\cdots(x_{k+2+\alpha}-x_{\gamma-1+\beta})(x_{k+2+\alpha}-x_{\gamma+\beta})}\nonumber\\
&+\cdots+\frac{(x_{k+1+k+1+\beta}
-x_{k+1+\beta})(x_{k+2+\alpha}-x_{k+2+\beta})\cdots(x_{k+2+\alpha}-x_{k+k+1+\beta})}{(x_{k+2+\alpha}-x_{1+\beta})\cdots(x_{k+2+\alpha}-x_{1+k+\beta})}
\nonumber \\
&+\frac{(x_{k+1+k+2+\beta}
-x_{k+2+\beta})(x_{k+2+\alpha}-x_{k+3+\beta})\cdots(x_{k+2+\alpha}-x_{k+k+2+\beta})}{(x_{k+2+\alpha}-x_{1+\beta})\cdots(x_{k+2+\alpha}-x_{1+k+\beta})}\nonumber\\
&+\cdots+\frac{(x_{k+1+\alpha}
-x_{1+\alpha})(x_{k+2+\alpha}-x_{1+\alpha})\cdots(x_{k+2+\alpha}-x_{k+\alpha})}
{(x_{k+2+\alpha}-x_{1+\beta})\cdots(x_{k+2+\alpha}-x_{1+k+\beta})}\nonumber\\
&+\frac{(x_{k+1+1+\alpha}
-x_{1+\alpha})(x_{k+2+\alpha}-x_{2+\alpha})\cdots(x_{k+2+\alpha}-x_{k+1+\alpha})}
{(x_{k+2+\alpha}-x_{1+\beta})\cdots(x_{k+2+\alpha}-x_{1+k+\beta})}.\nonumber
\end{align}
Now we divide first factor of the transition term, in the third line
above, into two halves by
$$x_{k+1+k+1+\beta}-x_{k+1+\beta}=(x_{k+2+\alpha}-x_{1+k+\beta})+(x_{k+1+k+1+\beta}-x_{k+2+\alpha}).$$
The first half triggers the recursive reduction on the first $k$ terms
exactly as in the first case, so the sum of the first $k$ terms equal
to $1$ and we get
\begin{align*}
B(A-1)=& -(x_{k+2+\alpha}-x_{k+k+2+\beta})
(x_{k+2+\alpha}-x_{k+2+\beta})\cdots(x_{k+2+\alpha}-x_{k+k+1+\beta})
\\
&+(x_{k+1+k+2+\beta}
-x_{k+2+\beta})(x_{k+2+\alpha}-x_{k+3+\beta})\cdots(x_{k+2+\alpha}-x_{k+k+2+\beta})\\
&+\cdots+(x_{k+1+\alpha}
-x_{1+\alpha})(x_{k+2+\alpha}-x_{1+\alpha})\cdots(x_{k+2+\alpha}-x_{k+\alpha})\\
&+(x_{k+1+1+\alpha}
-x_{1+\alpha})(x_{k+2+\alpha}-x_{2+\alpha})\cdots(x_{k+2+\alpha}-x_{k+1+\alpha}).
\end{align*}
Now we can do a recursive reduction starting from the first two terms,
the sum of which is
\begin{align*}
&\Big[x_{k+1+k+2+\beta}
-x_{k+2+\beta} - (x_{k+2+\alpha}-x_{k+2+\beta})\Big]
(x_{k+2+\alpha}-x_{k+3+\beta})\cdots(x_{k+2+\alpha}-x_{k+k+2+\beta}) \\
=&-(x_{k+2+\alpha}-x_{k+1+k+2+\beta})(x_{k+2+\alpha}-x_{k+3+\beta})
\cdots(x_{k+2+\alpha}-x_{k+k+2+\beta})
\end{align*}
This can be combined with the third term in a similar fashion and the
recursion continues. At the end, we get
\begin{align*}
B(A-1)=&-(x_{k+2+\alpha}-x_{k+1+\alpha})(x_{k+2+\alpha}-x_{1+\alpha})\cdots(x_{k+2+\alpha}-x_{k+\alpha})\\
&+(x_{k+1+1+\alpha}
-x_{1+\alpha})(x_{k+2+\alpha}-x_{2+\alpha})\cdots(x_{k+2+\alpha}-x_{k+1+\alpha})\\
=&\Big[x_{k+1+1+\alpha}
-x_{1+\alpha}- (x_{k+2+\alpha}-x_{1+\alpha})\Big]
(x_{k+2+\alpha}-x_{2+\alpha})\cdots(x_{k+2+\alpha}-x_{k+1+\alpha})=0.
\end{align*}
That is, we have shown that $A=1$.

With $A=1$ proved between these two cases, we have completed the
inductive argument, and hence the proof of the lemma.


\subsection{Proof of Lemma \ref{lem:hinv} (inverse representation)}

We prove Lemma~\ref{lem:hinv}, which claims that he
inverse of falling factorial basis matrix is
\begin{equation}
\label{A.eq:invH}
(H^{(k)})^{-1} = \left[\begin{array}{c} C \\
\frac{1}{k!} \cdot D^{(k+1)}
\end{array}\right],
\end{equation}
where $D^{(k+1)}$ is the $(k+1)^{st}$ order discrete difference
operator defined in \eqref{eq:dk}, and the rows of the matrix $C \in
\R^{(k+1)\times n}$ obey $C_1 = e_1$ and
\begin{equation*}
C_{i+1} =
\left[\frac{1}{i!} \cdot
(\Delta^{(i)})^{-1} \cdot D^{(i)}\right]_1,
\;\;\; i=1,\ldots k.
\end{equation*}
Again we use induction on $k$.  When $k=0$, it is easily verified that
$$
(H^{(0)})^{-1} = L_n^{-1}=
\left[\begin{array}{c}
                          e_1 \\
                          D^{(1)} \\
\end{array}\right] =
\left[\begin{array}{c}
                          e_1 \\
                          \frac{1}{0!}\cdot D^{(1)} \\
\end{array}\right].
$$
The rest of the inductive proof is relatively straightforward,
following from Lemma \ref{lem:recursive}, i.e., from
\eqref{A.eq:hk}. Inverting both sides of \eqref{A.eq:hk} gives
\begin{align}
  (H^{(k)})^{-1} &= \left[\begin{array}{cc}
I_k & 0 \\ 0 & L_{n-k}
\end{array}\right]^{-1}\cdot\left[\begin{array}{cc}
I_k & 0 \\ 0 & \Delta^{(k)}
\end{array}\right]^{-1}\cdot(H^{(k-1)})^{-1}\nonumber\\
&=\left[\begin{array}{cc}
I_{k} & 0 \\ 0 & L_{n-k}^{-1}
\end{array}\right]\cdot \left[\begin{array}{cc}
I_{k} & 0 \\ 0 & (\Delta^{(k)})^{-1}
\end{array}\right]\cdot (H^{(k-1)})^{-1}.
\nonumber 
\end{align}
Now, using that $L_{n-k}^{-1}=\begin{bmatrix} e_1 \\
                D^{(1)} \\
              \end{bmatrix}$,
and assuming that $(H^{(k-1)})^{-1}$ obeys \eqref{A.eq:invH},
\begin{align*}
  (H^{(k)})^{-1} &=\left[\begin{array}{cc}
I_{k} & 0 \\
0 & \left[\begin{array}{c}
                e_1 \\
                D^{(1)} \\
              \end{array}\right]
\end{array}\right]
\cdot
\left[\begin{array}{cc}
I_{k} & 0 \\ 0 & (\Delta^{(k)})^{-1}
\end{array}\right]
\cdot
\left[\begin{array}{c}
     e_1 \\
     \left[ \frac{1}{1!}(\Delta^{(1)})^{-1}D^{(1)} \right]_1 \\
     \vdots \\
     \left[ \frac{1}{(k-1)!}(\Delta^{(k-1)})^{-1}D^{(k-1)} \right]_1 \\
     \frac{1}{(k-1)!}\cdot D^{(k)}
              \end{array}\right] \\
   &=\left[\begin{array}{c}
     e_1 \\
     \left[ \frac{1}{1!}(\Delta^{(1)})^{-1}D^{(1)} \right]_1 \\
     \vdots \\
     \left[ \frac{1}{(k-1)!}(\Delta^{(k-1)})^{-1}D^{(k-1)} \right]_1 \\
     \frac{1}{k!}\left[\begin{array}{c}
                       e_1  \\
                       D^{(1)} \\
\end{array}\right]
     \cdot k(\Delta^{(k)})^{-1}\cdot D^{(k)} \\
\end{array}\right] = \left[\begin{array}{c}
     e_1 \\
     \left[ \frac{1}{1!}(\Delta^{(1)})^{-1}D^{(1)} \right]_1 \\
     \vdots \\
     \left[ \frac{1}{(k-1)!}(\Delta^{(k-1)})^{-1}D^{(k-1)} \right]_1 \\
     \left[ \frac{1}{(k)!}(\Delta^{(k)})^{-1}D^{(k)} \right]_1 \\
     \frac{1}{k!} \cdot D^{(k+1)}
\end{array}\right] =  \left[\begin{array}{c}
                    C \\
                    \frac{1}{k!} \cdot D^{(k+1)} \\
\end{array}\right],
\end{align*}
as desired.

\subsection{Algorithms for multiplication by $(H^{(k)})^T$ and
  $[(H^{(k)})^T]^{-1}$}
\label{app:algs}

Recall that, given a vector $y$, we write $y_{a:b}$ to denote its
subvector $(y_a,y_{a+1}, \ldots y_b)$, and we write $\mathrm{cumsum}$
and $\mathrm{diff}$ for the cumulative sum pairwise difference
operators.  Furthermore, we define
 $\mathrm{flip}$ to be the operator the reverses the order of its
 input, e.g., $\mathrm{flip}((1,2,3))=(3,2,1)$, and we write $\circ$
 to denote operator composition, e.g., $\mathrm{flip} \circ
 \mathrm{cumsum}$.  The remaining two
 algorithms from Lemma \ref{lem:fastcomp} are
 given below, in Algorithms~\ref{alg:hkt}~and~\ref{alg:hktinv}.

\setcounter{algorithm}{2}

\begin{algorithm}[tbh]
\caption{Multiplication by $(H^{(k)})^T$}
\label{alg:hkt}
\begin{algorithmic}
\STATE {\bfseries Input:} Vector to be multiplied $y \in \R^n$,
order $k \geq 0$, sorted inputs vector $x\in \R^n$.
\STATE {\bfseries Output:} $y$ is overwritten by $(H^{(k)})^{T} y$.
\FOR{$i=0$ to $k$}
\IF{$i\neq 0$}
\STATE $y_{(i+1):n}= y_{(i+1):n} \,./\, (x_{(i+1):n}-x_{1:(n-i)})$.
\ENDIF
\STATE
$y_{(i+1):n}=\mathrm{flip}
\circ\mathrm{cumsum}
\circ\mathrm{flip}(y_{(i+1):n})$.
\ENDFOR
\STATE Return $y$.
\end{algorithmic}
\end{algorithm}

\begin{algorithm}[tbh]
\caption{Multiplication by $[(H^{(k)})^T]^{-1}$}
\label{alg:hktinv}
\begin{algorithmic}
\STATE {\bfseries Input:} Vector to be multiplied $y \in \R^n$,
order $k \geq 0$, sorted inputs vector $x\in \R^n$.
\STATE {\bfseries Output:} $y$ is overwritten by $[(H^{(k)})^T]^{-1}y$.
\FOR{$i=k$ to $0$}
\STATE 
$y_{(i+1):n-1}=\mathrm{flip}
\circ\mathrm{diff}
\circ\mathrm{flip}(y_{(i+1):n})$.
\IF{$i\neq 0$}
\STATE $y_{(i+1):n}=(x_{(i+1):n}-x_{1:(n-i)})^{-1}
\, .\hspace{-2pt} * \,
 y_{(i+1):n}$.
\ENDIF
\ENDFOR

\STATE Return $y$.
\end{algorithmic}
\end{algorithm}

\subsection{Proof of Lemma~\ref{lem:hgdiff} (proximity to truncated
  power basis)}

Recall that we denote
$$
\delta = \max_{i=1,\ldots n} \, (x_i-x_{i-1}),
$$
and write $x_0=0$ for notational convenience.
Taking the elementwise difference between the falling factorial and
truncated power basis matrices, we get
\begin{equation}
\label{eq:diff}
H_{ij}-G_{ij}=
\begin{cases}
0 & \text{for}\;\, i=1,\ldots n,\; j=1 \\
\prod_{\ell=1}^{j-1}(x_i-x_\ell)-x_i^{j-1}
& \text{for}\;\, i>j-1,\;j=2,\ldots k+1\\
-x_i^{j-1}
& \text{for}\;\, i\leq j-1, \; j=2,\ldots k+1\\
0 & \text{for}\;\, i\leq j - \lceil k/2\rceil,
\; j\geq k+2 \\
-(x_i-x_{j-\lceil k/2 \rceil})^k
& \text{for}\;\, j- \lceil k/2 \rceil < i\leq j-1,
\; j\geq k+2 \\
\prod_{\ell=1}^{k} (x_i-x_{j-k-1+\ell})-(x_i-x_{j-\lceil
  k/2\rceil})^k&\text{for}\;\, i> j-1,\; j\geq k+2.
\end{cases}
\end{equation}
In the above, we use $\lceil z \rceil$ to denote the least
integer greater than or equal to $z$ (the ceiling function).
We will bound the absolute value of each nonzero difference
$H_{ij}-G_{ij}$ in \eqref{eq:diff}.  Starting with the second row,
\begin{align*}
  \left|\prod_{\ell=1}^{j-1}(x_i-x_\ell)-x_i^{j-1}\right|
&\leq x_i^{j-1}-(x_i-x_{j-1})^{j-1} \\
&=  x_{j-1}\left[x_i^{j-2}+x_i^{j-3}(x_i-x_{j-1})+\ldots+
    x_i(x_i-x_{j-1})^{j-3}+(x_i-x_{j-1})^{j-2} \right ] \\
& \leq  x_{j-1} \cdot (j-1) \cdot x_i^{j-2}
\leq k\delta \cdot k \cdot 1 \leq k^2 \delta.
\end{align*}
In the second line above, we used the expansion
\begin{equation}
\label{eq:exp}
a^k-b^k=(a-b)(a^{k-1}+a^{k-2}b+\ldots+b^{k-1}),
\end{equation}
and in the third line, we used the fact that $j-1 \leq k$, so that
$x_{j-1} \leq k\delta$, and also $0 \leq x_i \leq 1$.
The third row of \eqref{eq:diff} is simpler. Since $0 \leq x_i \leq 1$
and $i \leq j-1 < k$,
$$
|-x_i^{j-1}| \leq x_i \leq k\delta.
$$
For the fourth row in \eqref{eq:diff}, using the range of $i,j$,
and the fact that $k\delta \leq 1$, 
$$
|-(x_i-x_{j-\lceil k/2\rceil})^k| 
\leq (x_{j-1}-x_{j-\lceil k/2\rceil})^k \leq
(k\delta)^k \leq k\delta.
$$
This leaves us to deal with the last row in \eqref{eq:diff}.
Defining $p=i$, $q=j-(k+1)$, the problem transforms into bounding
\begin{equation*}
\prod_{\ell=1}^{k} (x_p-x_{\ell+q}) -
(x_p-x_{\lfloor\frac{k+2}{2}\rfloor+q})^k,
\end{equation*}
for any $p=k+2,k+3,\ldots n$, $q=1,\ldots p-k$, where now $\lfloor z
\rfloor$ denotes the greatest integer less than or equal to $z$ (the
floor function).
We let \smash{$\mu_{pq}=x_p-x_{\lfloor\frac{k+2}{2}\rfloor+q}$} and
\smash{$\eta_q=x_p-x_{q+1} -\mu_{pq}$}. Note that $\eta_q$ is the gap
between the maximum multiplicant in the first term above and
$\mu_{pq}$.  Then
$$
\eta_q=x_{\lfloor\frac{k+2}{2}\rfloor+q}-x_{q+1} \leq k\delta.
$$
Therefore
\begin{align*}
\prod_{\ell=1}^{k}(x_p-x_{\ell+q})-
(x_p-x_{\lfloor \frac{k+2}{2} \rfloor+q})^k
&\leq (x_p-x_{1+q})^k - \mu_{pq}^k \\
&= (\mu_{pq}+\eta_q)^k-\mu_{pq}^k \\
&=k \delta \cdot \sum_{\ell=0}^{k-1}
(\mu_{pq}+\eta_q)^\ell\mu_{pq}^{k-\ell} \\
&\leq k^2 \delta \cdot (\mu_{pq}+\eta_q)^k
\leq k^2 \delta.
\end{align*}
The third line above follows again from the expansion \eqref{eq:exp},
and the fact that
$\eta_q \leq k\delta$.  The fourth line uses $\mu_{pq}+\eta_q\geq
\mu_{pq}$, and ultimately $\mu_{pq}+\eta_q = x_p-x_{1+q} \in
[0,1]$.  This completes the proof.

\subsection{Proof of Theorem~\ref{thm:fixedconv} (trend filtering
  rate, fixed inputs)}

This proof follows the same strategy as the convergence proofs in
\citet{trendfilter}.
Recall that the trend filtering estimate \eqref{eq:tf} can be
expressed in terms of the lasso problem \eqref{eq:tfh}, in that
\smash{$\hbeta=H^{(k)}\halpha$}; also consider
consider the problem
\begin{equation}
\label{eq:lassog}
\hat{\theta} = \argmin_{\theta \in \R^n} \,
\half \|y-G^{(k)} \theta\|_2^2 +
\lambda' \cdot \hspace{-3pt} \sum_{j=k+2}^n |\theta_j|,
\end{equation}
where $G^{(k)}$ is the truncated power basis matrix of order $k$.
Let $\mu=(f_0(x_1),\ldots f_0(x_n)) \in \R^n$ denote the true function
evaluated across the inputs.  Then under the assumptions of Theorem
\ref{thm:fixedconv}, it is known that
\begin{equation*}
\|G^{(k)}\hat{\theta}-\mu\|_2^2 = O_\P(n^{-(2k+2)/(2k+3)}),
\end{equation*}
when $\lambda=\Theta(n^{1/(2k+3)})$; see Theorem 10 of
\citet{locadapt}.  It now suffices to show that
\smash{$\|H^{(k)}\halpha - G^{(k)}\hat{\theta}\|_2^2 =
O_\P(n^{-(2k+2)/(2k+3)})$}, since
\smash{$\|H^{(k)}\halpha - \mu\|_2^2 \leq
2\|H^{(k)}\halpha - G^{(k)}\hat{\theta}\|_2^2 +
2\|G^{(k)}\hat{\theta}-\mu\|_2^2$}.
For this, we can use the results in
Appendix B of \citet{trendfilter}, specifically Corollary 4 of this
work, to argue that we have
\smash{$\|H^{(k)}\halpha - G^{(k)}\hat{\theta}\|_2^2 =
O_\P(n^{-(2k+2)/(2k+3)})$} as long as
$\lambda=(1+\delta)\lambda'$ for any $\delta>0$, and
\begin{equation*}
n^{(2k+2)/(2k+3)} \cdot \max_{i,j=1,\ldots n} \,
|G_{ij}^{(k)}-H_{ij}^{(k)}| \rightarrow 0 \;\;\; \text{as}\;\,
  n\rightarrow \infty.
\end{equation*}
But by Lemma \ref{lem:hgdiff}, and our condition \eqref{eq:maxgap} on
the inputs, we have
\smash{$\max_{i,j=1,\ldots n} |G_{ij}^{(k)}-H_{ij}^{(k)}| \leq k^2
\log{n}/n$}, which verifies the above, and hence gives the result.

\subsection{Proof of Lemma \ref{lem:randgap} (maximum gap between
  random inputs)}

Given sorted i.i.d.\ draws $x_1 \leq \ldots \leq x_n$ from a
continuous distribution supported on $[0,1]$, whose density is bounded
below by $p_0>0$, we consider the maximum gap
$\delta=\max_{i=1,\ldots n} (x_i-x_{i-1})$ (recall that we set
$x_0=0$ for notational convenience).  This is a well-studied
quantity.  In the case of a uniform distribution on $[0,1]$, we know
that the spacings vector follows a symmetric Dirichelet distribution,
which is equivalent to uniform sampling from an $n$-simplex, e.g., see
\citet{david1970order}.  Furthermore, the asymptotics of the $k$th
largest gap have also been extensively studied, e.g., in
\citet{barbe1992limiting}. Here, we provide a simple finite sample
bound on $\delta$, without using distributional or geometric
characterizations, but rather a direct argument based on binning.




Consider an arbitrary point $x$ in $[0,1-\alpha]$.  Then the
probability that at least one draw from our underlying distribution
occurs in $[x,x+\alpha]$ is bounded below by
$1-(1-p_0\alpha)^{n}$.  Now divide $[0,1]$ into bins of length
$\alpha$ (the last bin can be overlapping with the second to last
bin).  Note that the event in which there is at least one sample point in
each bin implies that the maximum gap $\delta$ between adjacent points
is less than or equal to $2\alpha$. By the union bound, this event
occurs with probability at least
$1-\lceil \frac{1}{\alpha}\rceil(1-p_0 \alpha)^{n}$.

Let $\alpha = r\log n/(p_0 n)$, and assume $n$ is sufficiently
large so that $r\log n/(p_0 n)<1$.  Then we have
\begin{align*}
\Big\lceil \frac{1}{\alpha}\Big\rceil
(1-p_0\alpha)^{n}
& \leq \Big(\frac{1}{\alpha}+1\Big)
(1-p_0\alpha)^{n} = \frac{p_0 n+r\log n}{r\log n}
\Big(1-\frac{r \log n}{n}\Big)^n \\
&\leq 2p_0 n \exp(-r\log n) = 2p_0 n^{1-r}.
\end{align*}
Plugging in $r=11$, we get the desired result for $C=22$, i.e., with
probability at least $1-2p_0n^{-10}$, the maximum gap satisfies
$\delta \leq 22 \log{n}/(p_0 n)$.

\subsection{Proof of Corollary~\ref{cor:randomconv} (trend filtering
  rate, random inputs)}

The proof of this result is entirely analogous to the proof of Theorem
\ref{thm:fixedconv}; the only difference is that
\begin{equation*}
\max_{i=1,\ldots n-1}\, (x_{i+1}-x_i) = O_\P(\log{n}/n),
\end{equation*}
(i.e., convergence in probability now), and so accordingly,
\begin{equation*}
n^{(2k+2)/(2k+3)} \cdot \max_{i,j=1,\ldots n} \,
|G_{ij}^{(k)}-H_{ij}^{(k)}| \;\;\overset{p}{\rightarrow} \;\; 0 \;\;\; \text{as}\;\,
  n\rightarrow \infty,
\end{equation*}
employing Lemmas \ref{lem:hgdiff} and \ref{lem:randgap}. The
same arguments now apply; the stability result in Corollary 4 in
Appendix B of \citet{trendfilter} must now be applied to random
predictor matrices, but this is an extension that is straightforward
to verify.

\section{The higher order KS test}
\label{app:ks}

\subsection{Motivating arguments}

As described in the text, the classical KS test is
\begin{equation}
\label{eq:origform}
\KS(X_{(m)},Y_{(n)}) = \max_{z_j \in Z_{(m+n)}} \,
\left|\frac{1}{m}\sum_{i=1}^m 1\{x_i \leq z_j\} -
\frac{1}{n}\sum_{i=1}^n 1\{y_i \leq z_j\} \right|,
\end{equation}
over samples $X_{(m)}=(x_1,\ldots x_m)$ and
$Y_{(n)}=(y_1,\ldots y_n)$, written in combined form as
$Z_{(m+n)} =X_{(m)}\cup Y_{(n)} = (z_1,\ldots z_{m+n})$.
It is well-known that the above definition is equivalent to
\begin{equation}
\label{eq:varform}
\KS(X_{(m)},Y_{(n)}) = \max_{f \,:\, \TV(f) \leq 1} \,
\left|\hat{\E}_{X_{(m)}}[f(X)] - \hat{\E}_{Y_{(n)}} [f(Y)]\right|,
\end{equation}
where we write \smash{$\hat{\E}_{X_{(m)}}$} for the empirical
expectation under $X_{(m)}$, so \smash{$\hat{\E}_{X_{(m)}} [f(X)] =
  1/m \sum_{i=1}^m f(x_i)$}, and similarly for
\smash{$\hat{\E}_{Y_{(n)}}$}.   The equivalence between these two
definitions follows from the fact that the maximum in
\eqref{eq:varform} always occurs at an indicator function,
 $f(x)=1\{x\leq z_i\}$, for some $i=1,\ldots m+n$.

We now will step through a sequence of motivating arguments that lead
to the definition of the higher order KS test in \eqref{eq:kskg}.  The
basic idea is to alter the constraint set in \eqref{eq:varform}, and
consider functions of bounded variation in their $k$th derivative, for
some fixed $k \geq 0$. This gives
\begin{equation}
\label{eq:varformk1}
\max_{f \,:\, \TV(f^{(k)}) \leq 1} \,
\left|\hat{\E}_{X_{(m)}}[f(X)] - \hat{\E}_{Y_{(n)}} [f(Y)]\right|.
\end{equation}
Is it possible to compute such a quantity?
By a variational result in \citet{locadapt}, the maximum in
\eqref{eq:varformk1} is always achieved by a $k$th order spline
function.  In principle, if we knew some finite set $T$ containing the
knots of the maximizing spline, then we could restrict our attention
to the space of splines with knots in $T$.  However, when $k \geq 2$,
such a set $T$ is not generically easy to find, because the knots of
the maximizing spline in \eqref{eq:varformk1} can lie outside of the
set of data samples $Z_{(m+n)}=\{z_1,\ldots z_{m+1}\}$
\citep{locadapt}.  Therefore, we further restrict the functions in
consideration in \eqref{eq:varformk1} to be $k$th order splines with
knots contained in $Z=Z_{(m+n)}$. Letting
\smash{$\cS^{(k)}_Z$} denote the space of such spline
functions, we hence examine
\begin{equation}
\label{eq:varformk2}
\max_{f \in \cS^{(k)}_Z \,:\, \TV(f^{(k)}) \leq 1} \,
\left|\hat{\E}_{X_{(m)}}[f(X)] - \hat{\E}_{Y_{(n)}} [f(Y)]\right|.
\end{equation}

As \smash{$\cS^{(k)}_Z$} is a finite-dimensional
function space (in fact, $(m+n)$-dimensional), we can rewrite
\eqref{eq:varformk2} in a parametric form, similar to
\eqref{eq:origform}.  Let $g_1,\ldots g_{m+n}$ denote the $k$th order
truncated power basis with knots over the set of joined data
samples $Z$.  Then any function \smash{$f \in \cS^{(k)}_Z$} with
\smash{$\TV(f^{(k)}) \leq 1$} can be expressed as
\smash{$f = \sum_{j=1}^{m+n} \alpha_j g_j$}, where the coefficients
satisfy \smash{$\sum_{j=k+2}^{m+n} |\alpha_j| \leq 1$}.  In terms of
the evaluations of the function $f$ over $z_1,\ldots z_{m+n}$, we have
$$
\big(f(z_1),\ldots f(z_{m+n})\big) = G^{(k)} \alpha,
$$
where \smash{$G^{(k)}$} is the truncated power basis matrix, i.e., its
columns give the evaluations of $g_1,\ldots g_{m+n}$ over the points
$z_1,\ldots z_{m+n}$. Therefore \eqref{eq:varformk2} can be
re-expressed as
\begin{equation}
\label{eq:primal1}
\max_{\sum_{j=k+2}^{m+n} |\alpha_j| \leq 1} \,
\left|\frac{1}{m}\mathds{1}_{X_{(m)}}^T G^{(k)} \alpha
  - \frac{1}{n}\mathds{1}_{Y_{(n)}}^TG^{(k)} \alpha\right|.
 \end{equation}
Here \smash{$\mathds{1}_{X_{(m)}}$} is an indicator vector of length
$m+n$, indicating the membership of each point in the joined sample
$Z_{(m+n)}$ to the set $X_{(m)}$.  The analogous definition is made
for \smash{$\mathds{1}_{Y_{(n)}}$}.

Upon inspection, some care must be taken in evaluating the maximum in
\eqref{eq:primal1}.  Let us decompose the coefficient vector into
blocks as $\alpha=(\alpha_1,\alpha_2)$, where $\alpha_1$ denotes the
first $k+1$ coefficients and $\alpha_2$ the last $m+n-k-1$.  Then the
constraint in \eqref{eq:primal1} is simply $\|\alpha_2\|_1 \leq 1$,
and  it is not hard to see that since $\alpha_1$ is unconstrained, we
can choose it to make the criterion in \eqref{eq:primal1} arbitrarily
large.   Therefore, in order to make \eqref{eq:primal1} well-defined
(finite), we employ the further restriction $\alpha_1=0$, yielding
\begin{equation}
\label{eq:primal2}
\max_{\|\alpha_2\|_1 \leq 1} \,
\left|\frac{1}{m}\mathds{1}_{X_{(m)}}^T G_2^{(k)} \alpha_2
  - \frac{1}{n}\mathds{1}_{Y_{(n)}}^TG_2^{(k)} \alpha_2\right|,
 \end{equation}
where \smash{$G_2^{(k)}$} denotes the last $m-n-k-1$ columns of
\smash{$G^{(k)}$}.  A simple duality argument shows that
\eqref{eq:primal2} can be written in terms of the $\ell_\infty$ norm,
finally giving
\begin{equation}
\label{eq:dualG}
\KS_G^{(k)}(X_{(m)},Y_{(n)}) =
\left\| (G_2^{(k)})^T \left(
\frac{ \mathds{1}_{X_{(m)}} }{m} -
\frac{ \mathds{1}_{Y_{(n)}} }{n} \right) \right\|_\infty,
\end{equation}
matching the our definition of the $k$th order KS test in
\eqref{eq:kskg}.   Note that when $k=0$, this reduces to the usual
(classic) KS test in \eqref{eq:origform}.

For $k \geq 1$, unlike the usual KS test which requires
$O(m+n)$ operations, the $k$th order KS test in \eqref{eq:dualG}
requires $O((m+n)^2)$ operations, due to the lower triangular nature
of \smash{$G^{(k)}$}.
Armed with our falling factorial basis, we can approximate
\smash{$\KS_G^{(k)}(X^m,Y^n)$} by
\begin{equation}
\label{eq:dualH}
\KS_H^{(k)}(X_{(m)},Y_{(n)}) =
\left\| (H_2^{(k)})^T \left(
\frac{ \mathds{1}_{X_{(m)}} }{m} -
\frac{ \mathds{1}_{Y_{(n)}} }{n} \right) \right\|_\infty,
\end{equation}
where \smash{$H^{(k)}$} is the $k$th order falling factorial basis
matrix (and  \smash{$H^{(k)}_2$} its last $m+n-k-1$ columns) over the
points $z_1,\ldots z_{m+n}$.  After sorting $z_1,\ldots z_{m+n}$, the
statistic in \eqref{eq:dualH} can be computed in $O(k(m+n))$ time;
see Algorithm \ref{alg:hkt}, described above in Section
\ref{app:algs}.

\subsection{Additional experiments}
\begin{figure}[h]
  \centering
  \subfigure[Experiment 1]{
    \includegraphics[width=0.31\linewidth]{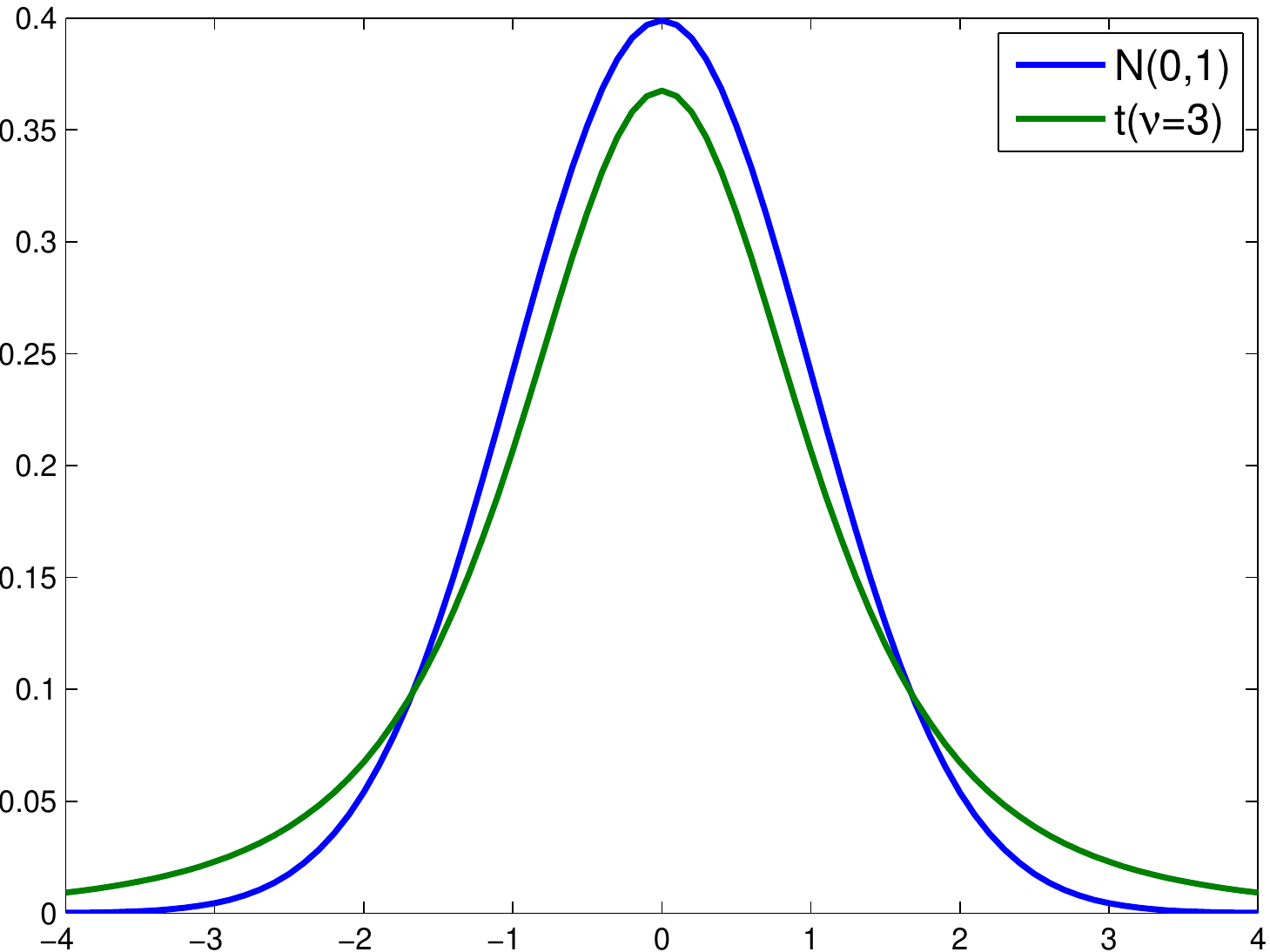}
  }
  \subfigure[Experiment 2]{
    \includegraphics[width=0.31\linewidth]{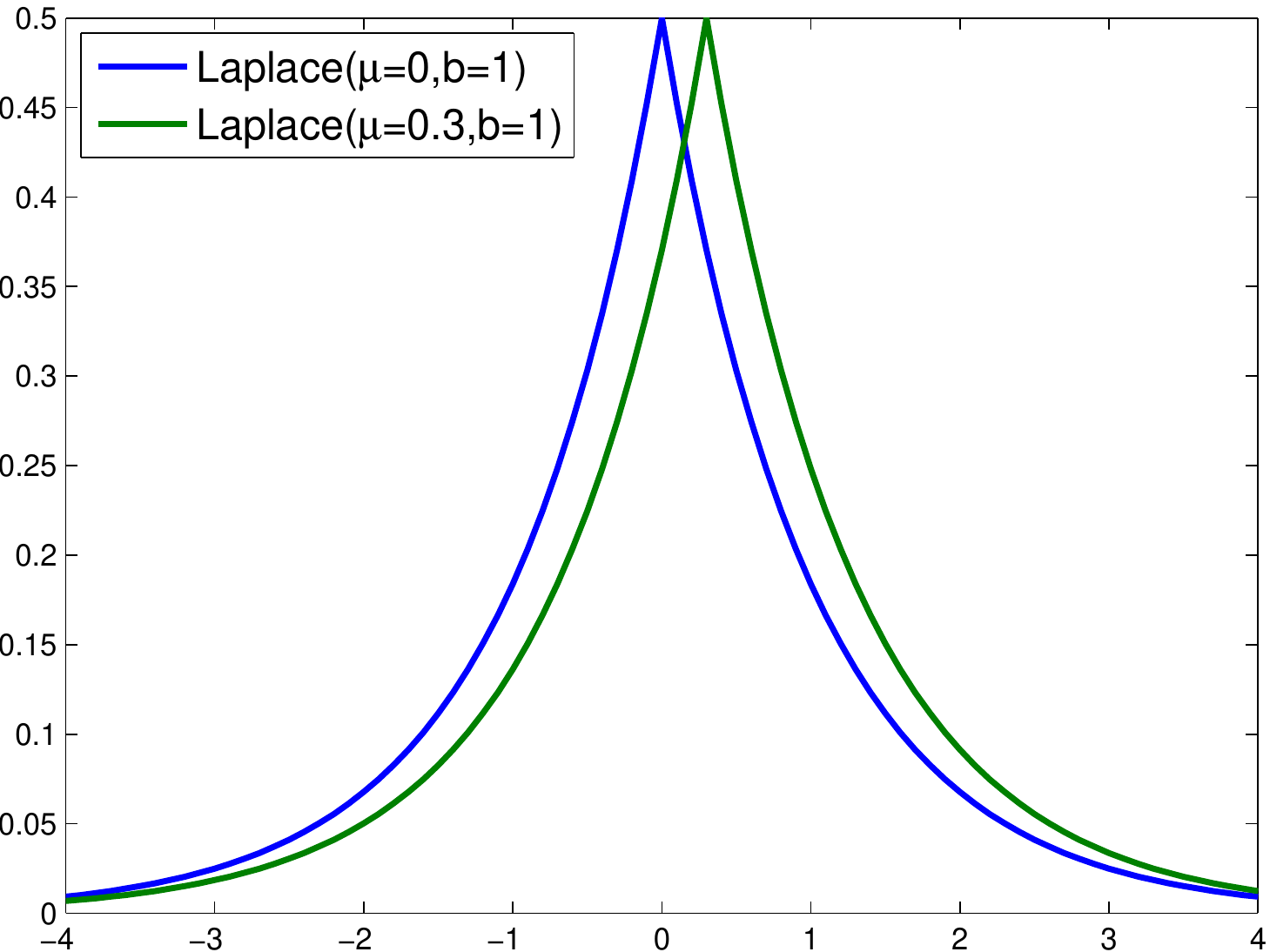}
  }
  \subfigure[Experiment 3]{
    \includegraphics[width=0.31\linewidth]{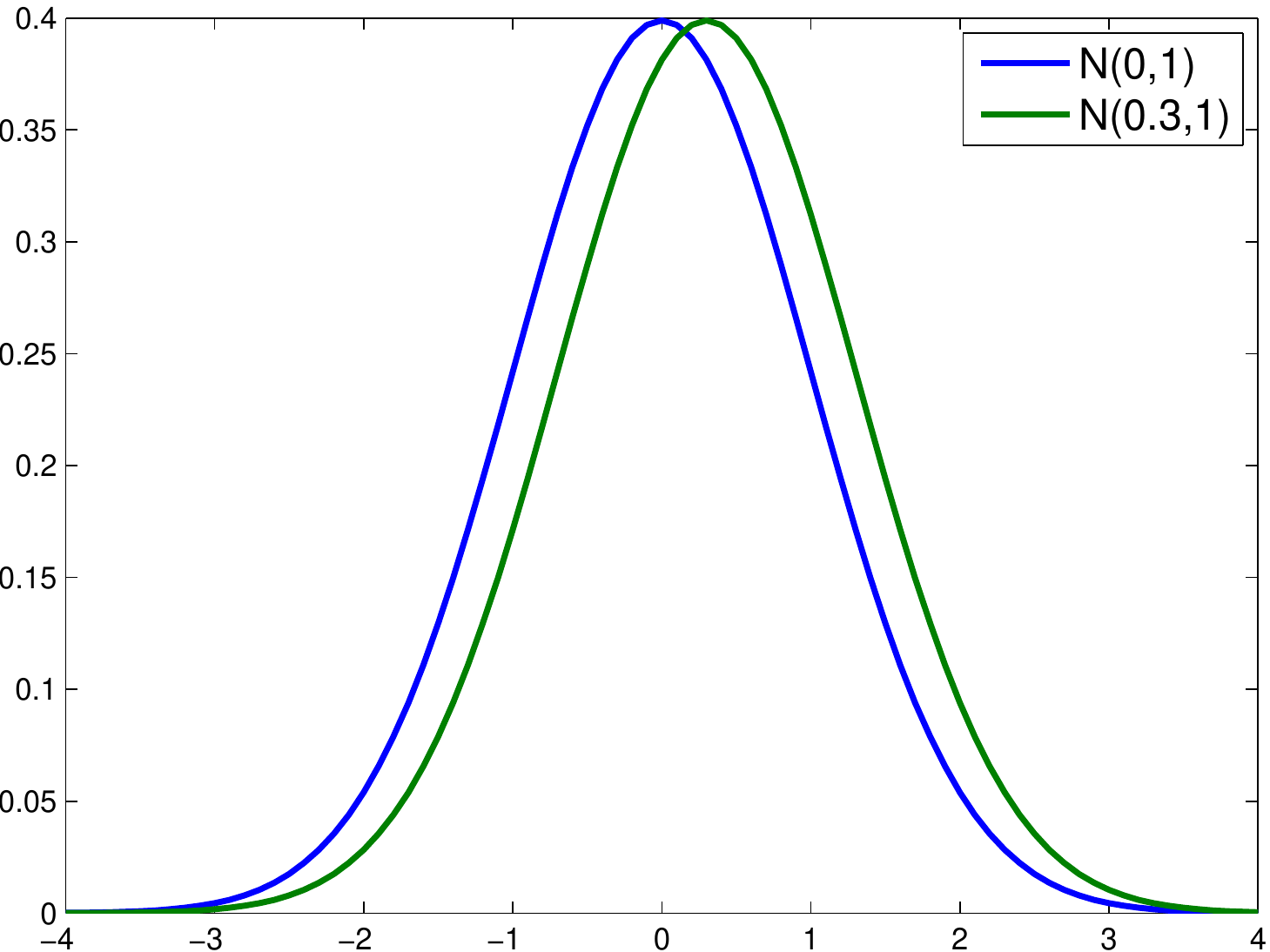}
  }
  \caption{An illustration of distribution $P$ vs. $Q$ in our numerical
    experiments.}
  \label{fig:illus}
\end{figure}

In the main text, we presented two numerical experiments, on testing
between samples from different distributions $P,Q$.  In the first
experiment $P=N(0,1)$ and $Q=t_3$, so the difference between $P,Q$
was mainly in the tails; in the second, $P=\mathrm{Laplace}(0)$ and
$Q=\mathrm{Laplace}(0.3)$, and the difference between $P,Q$ was
mainly in the centers of the distributions.  The first experiment
demonstrated that the power of the higher order KS test generally
increased as we increased the polynomial degree $k$, the second
demonstrated the opposite, i.e., that its power generally decreased
for increasing $k$.  Refer back to Figures \ref{fig:exp1} and
\ref{fig:exp2} in the main text.

We should note that the first experiment was not carefully crafted in
any way; the same performance is seen with a number of similar setups.
However, we did have to look carefully to reveal the negative behavior
shown in the second experiment.  For example, in detecting the
difference between mean-shifted standard normals (as opposed to
Laplace distributions), the higher order KS tests do not encounter
nearly as much difficulty.  To demonstrate this, we examine a third
experiment here with $P=N(0,1)$ and $Q=N(0.3,1)$.
Figure~\ref{fig:illus} gives a visual illustration of the
distributions across the three experimental setups (the first two
considered in the main text, and the third investigated here).

\begin{figure}[tb!]
\centering
\subfigure[Comparing higher order KS tests]{
  \includegraphics[width=0.48\linewidth]{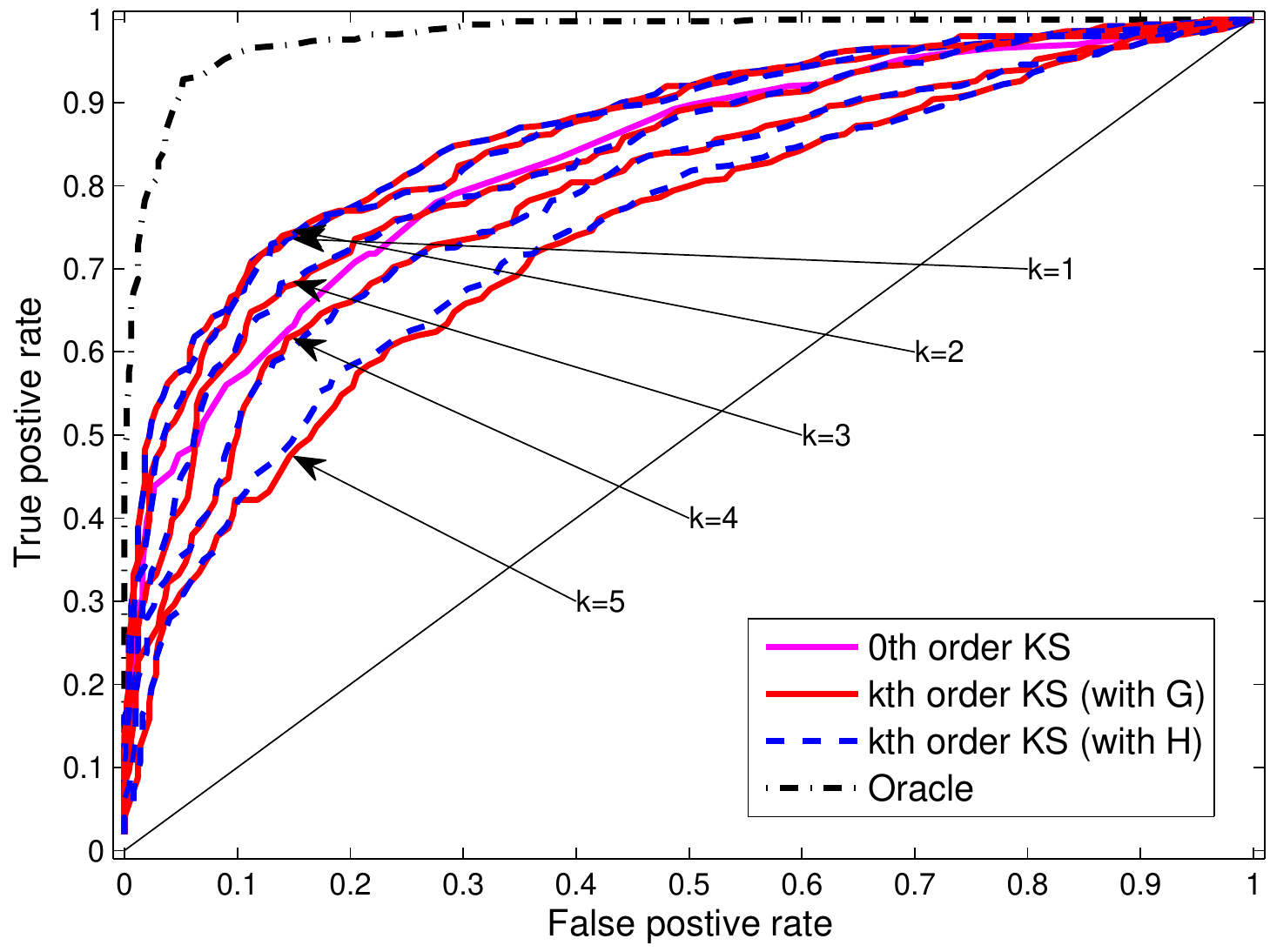}
}
\subfigure[Comparing other tests]{
  \includegraphics[width=0.48\linewidth]{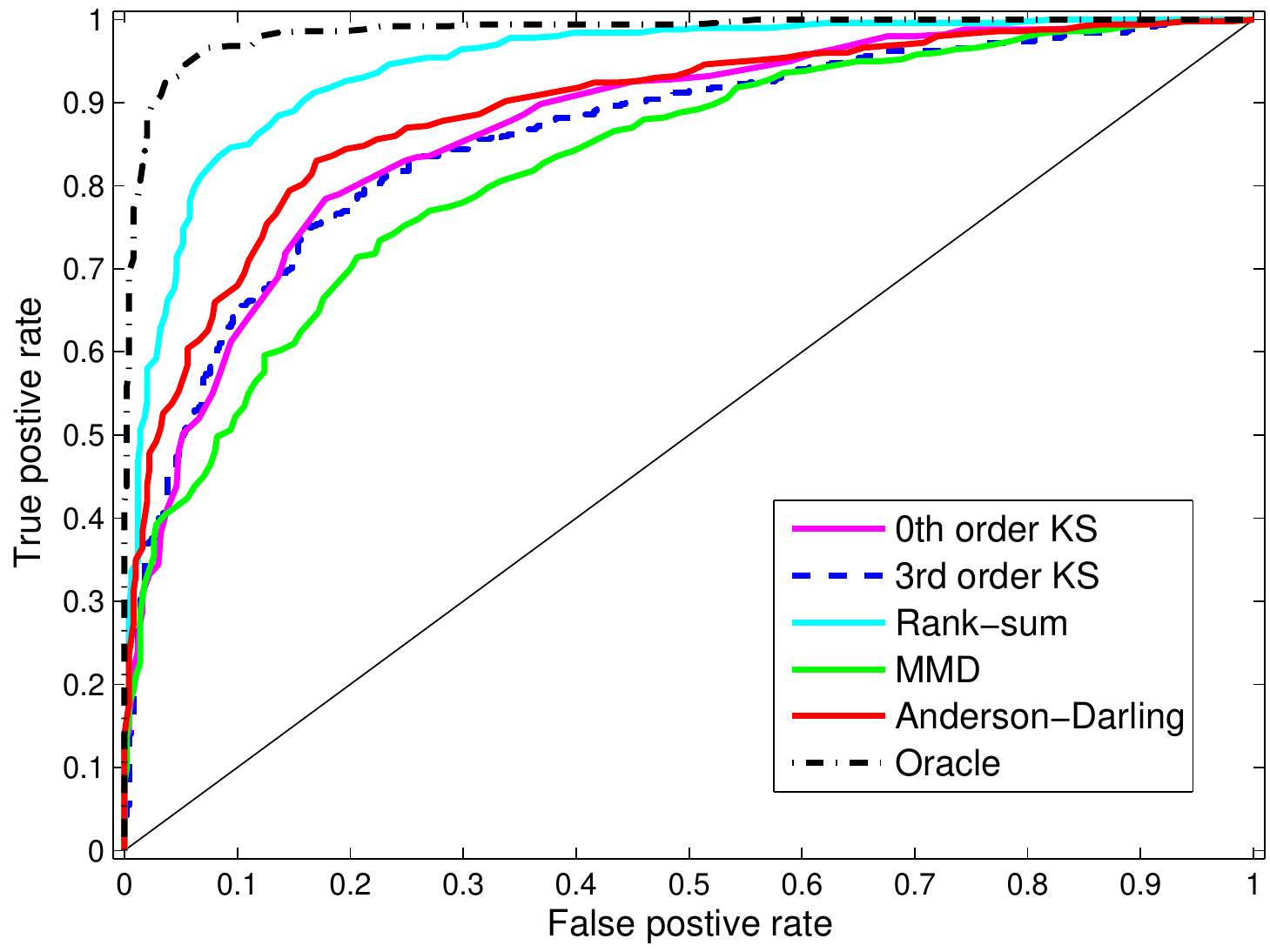}
}
\caption{ROC curves for experiment 3, normal vs.\ shifted normal.}
\label{fig:exp3}
\end{figure}

The ROC curves for experiment 3 are given in
Figure~\ref{fig:exp3}. The left panel shows that the test for $k=1$
improves on the usual test ($k=0$), even though the difference between
the two distributions is mainly near their centers.  The right panel
shows that the higher order KS tests are competitive
with other commonly used nonparametric tests in this setting.
The results of this experiment hence suggest that the higher order KS tests
provide a utility beyond simply detecting finer tail
differences, and the tradeoff induced by varying the polynomial order
$k$ is not completely explained as a tradeoff between tail and center
sensitivity.

\begin{figure}[tbh]
  \centering
  \subfigure[Comparing higher order KS tests]{
    \includegraphics[width=0.48\linewidth]{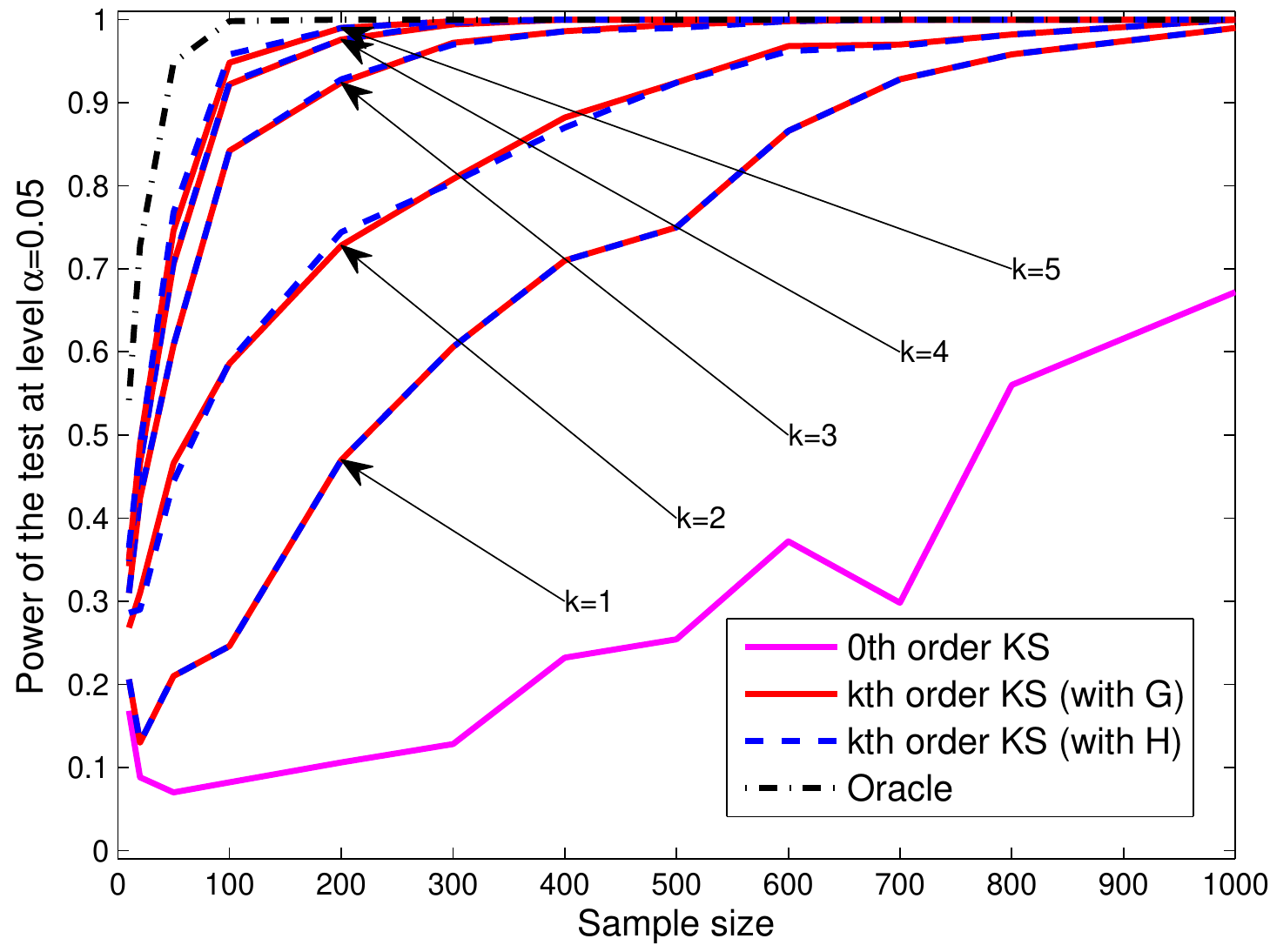}
  }
  \subfigure[Comparing other tests]{
    \includegraphics[width=0.48\linewidth]{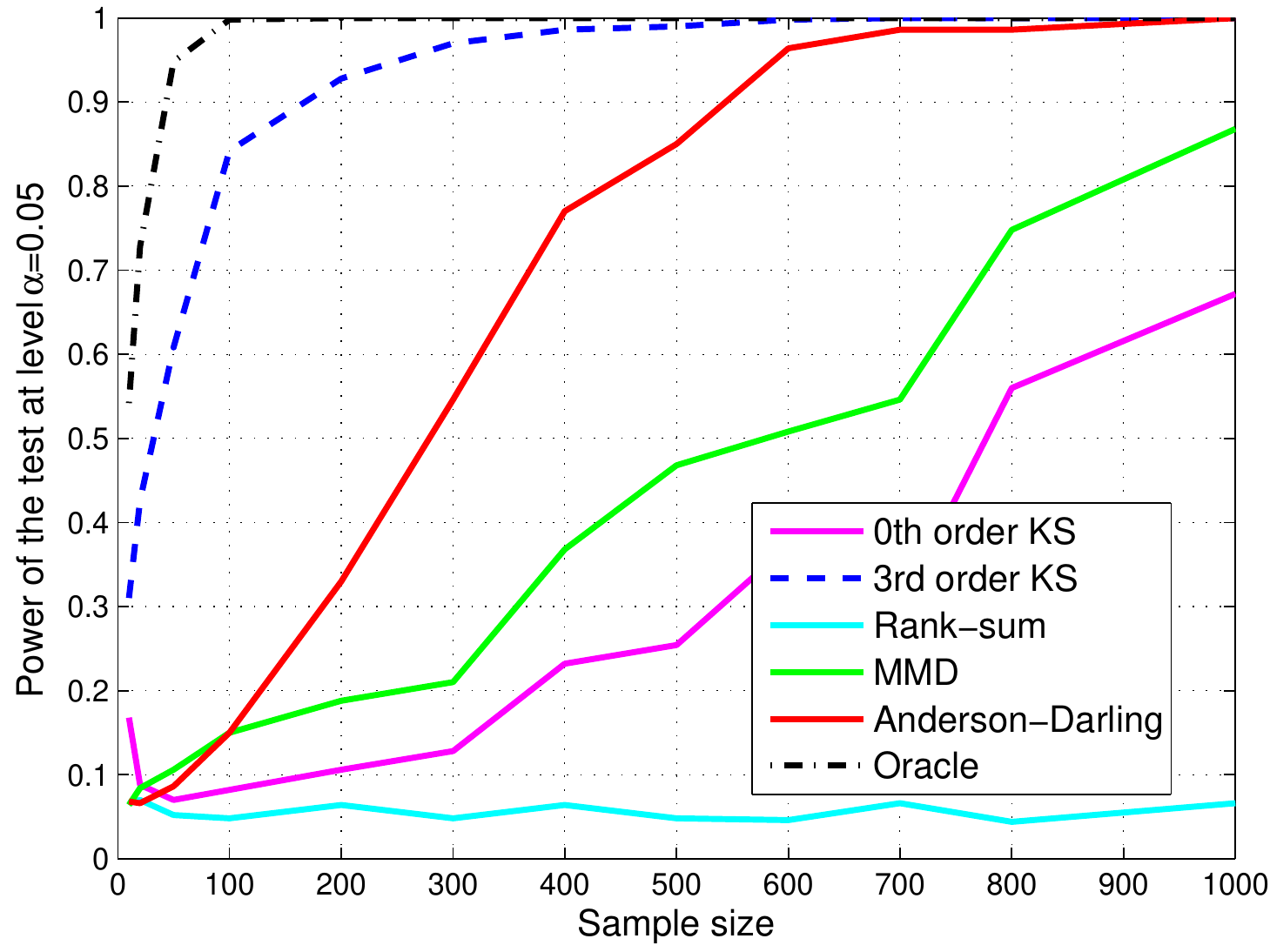}
  }
  \caption{Sample complexities at the level $\alpha=0.05$ for experiment
    1, normal vs.\ t.}
  \label{fig:exp1samp}
\end{figure}

\begin{figure}[tbh]
  \centering
  \subfigure[Comparing higher order KS tests]{
    \includegraphics[width=0.48\linewidth]{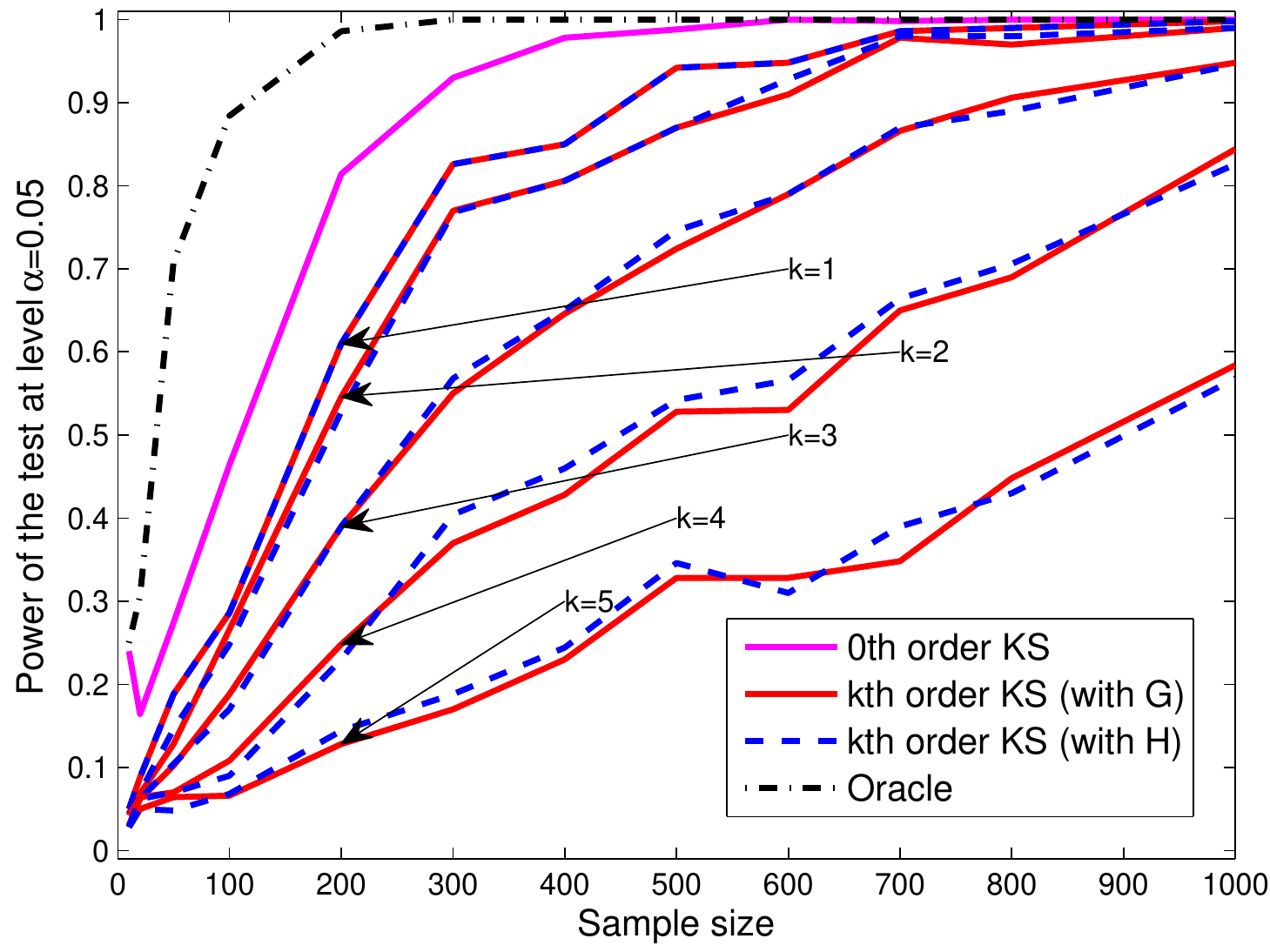}
  }
  \subfigure[Comparing other tests]{
    \includegraphics[width=0.48\linewidth]{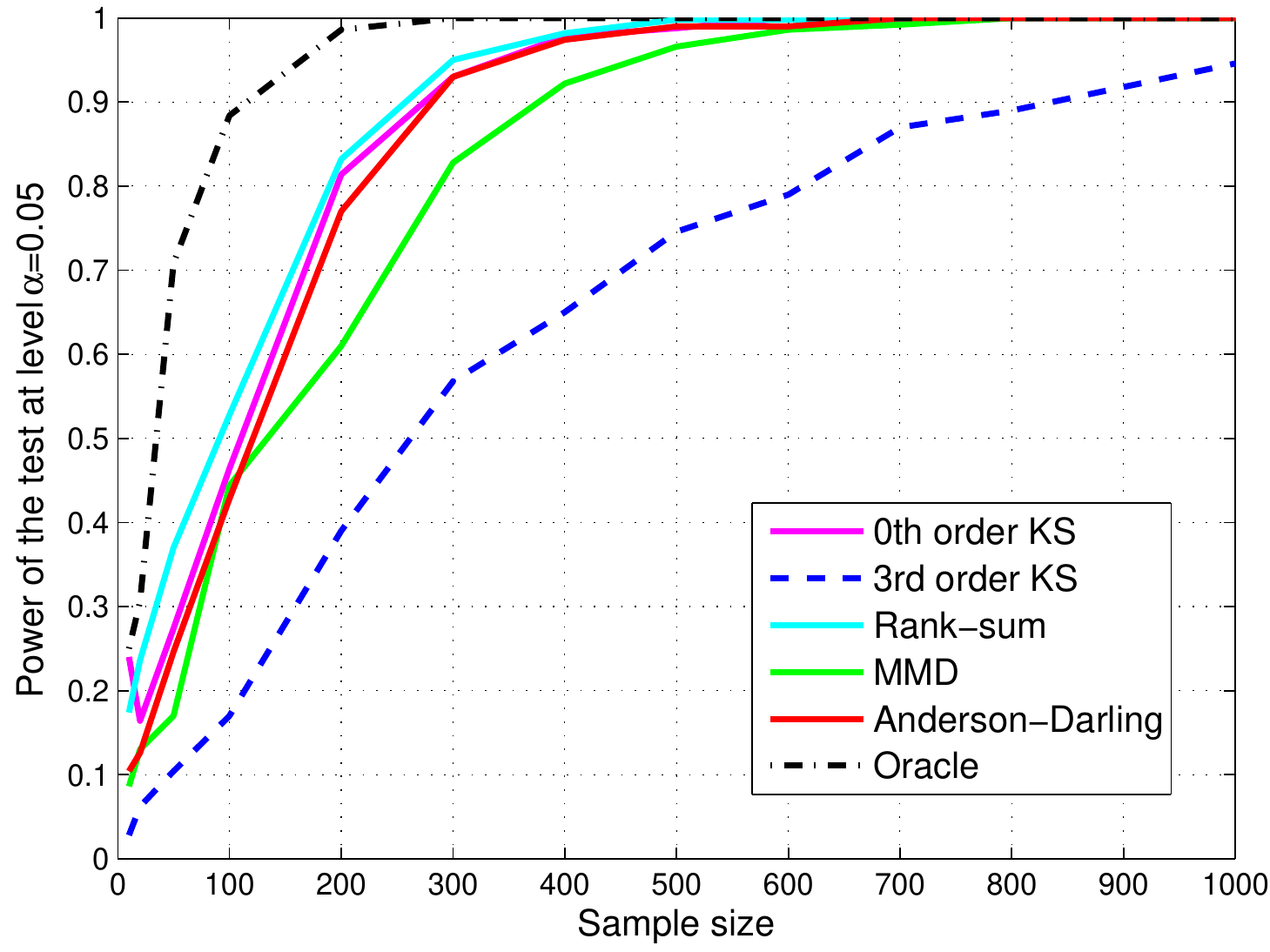}
  }
  \caption{Sample complexities at level $\alpha=0.05$ in experiment 2,
    Laplace vs.\ shifted Laplace.}
  \label{fig:exp2samp}
\end{figure}

\begin{figure}[tbh]
  \centering
  \subfigure[Comparing higher order KS tests]{
    \includegraphics[width=0.47\linewidth]{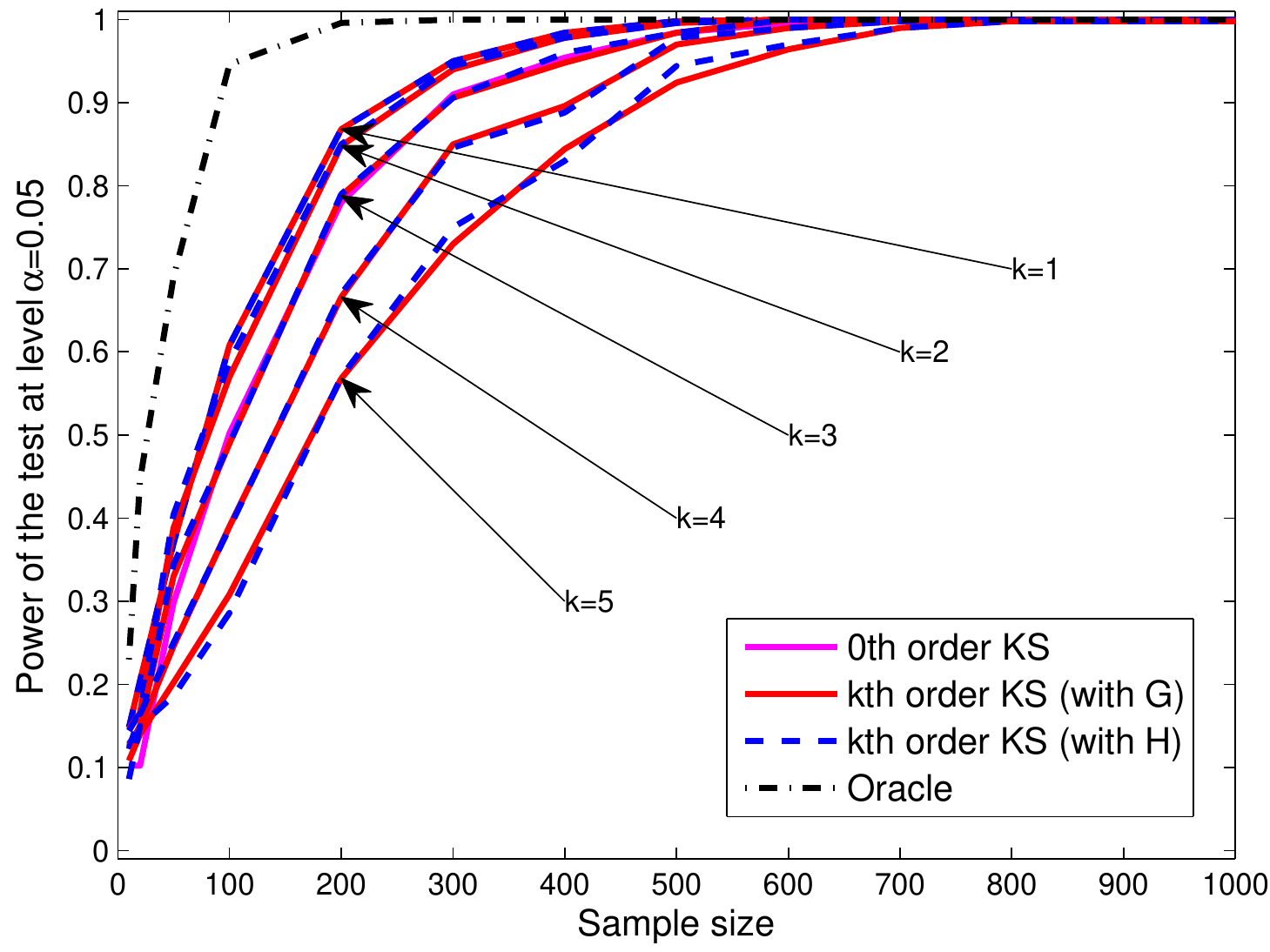}
  }
  \subfigure[Comparing other tests]{
    \includegraphics[width=0.47\linewidth]{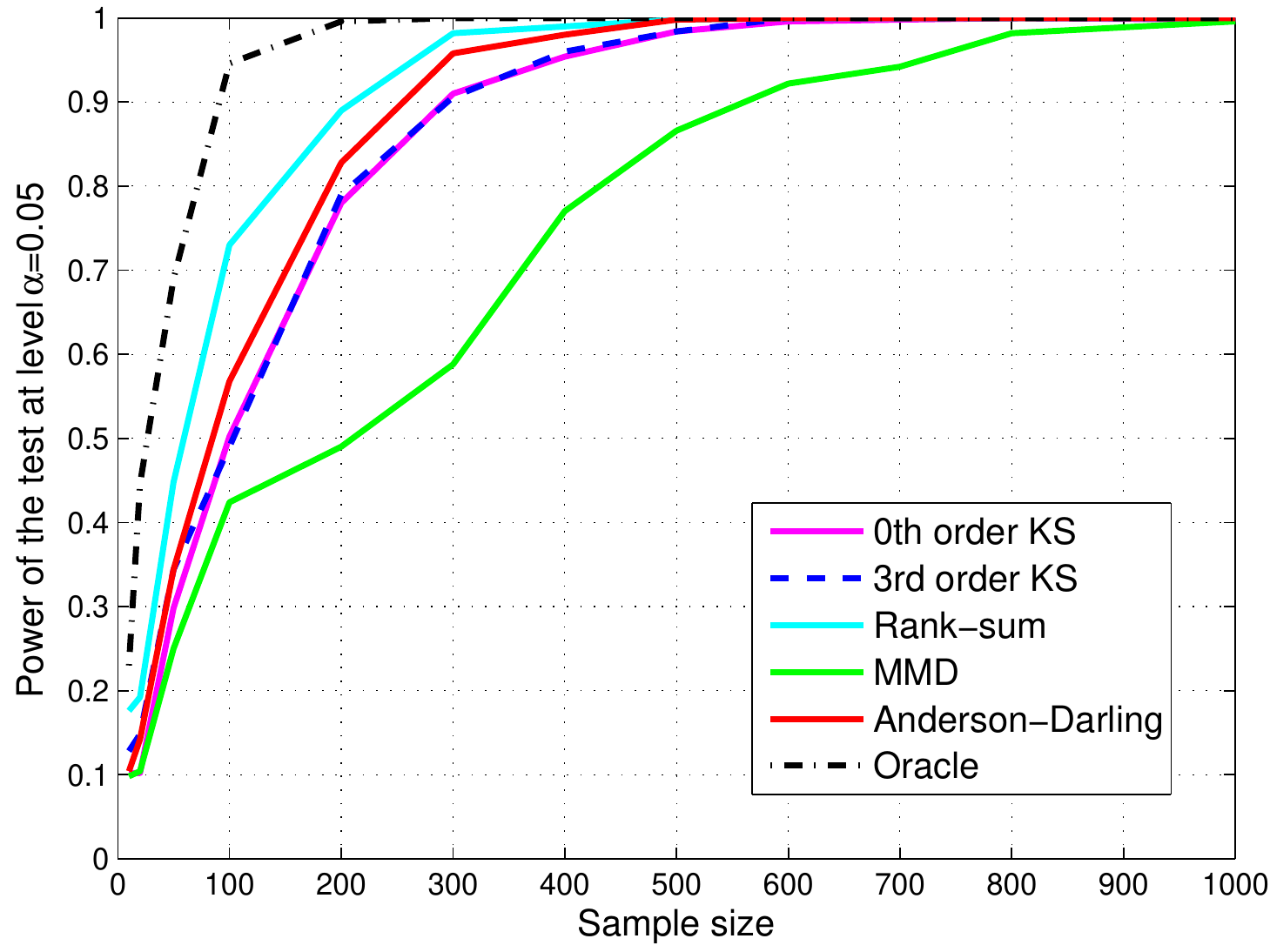}
  }
  \caption{Sample complexities at level $\alpha=0.05$ in experiment 3,
    normal vs.\ shifted normal.}
  \label{fig:exp3samp}
\end{figure}

We also study the sample complexity of tests in the three experimental
setups.  Specifically, over $R=1000$ repetitions, we find the true
positive rate associated with a 0.05 false positive rate, as we let
$n$ vary over $10,20,50,100,200,\ldots 1000$.  The results for this
sample complexity sudy are shown in Figures \ref{fig:exp1samp},
\ref{fig:exp2samp}, and \ref{fig:exp3samp}.  We see that the
higher order KS tests perform quite favorably the first experimental
setup, not so favorably in the second, and somewhere in the
middle in the third.

\bibliography{ryantibs}
\bibliographystyle{icml2014}

\end{document}